%% file: main.tex
\documentclass[acmlarge]{acmart}
\setcitestyle{super,sort&compress}
\usepackage{booktabs} 
\usepackage{graphicx}
\usepackage{amsmath}
\usepackage{lscape}
\usepackage{multirow}
\usepackage[ruled,linesnumbered]{algorithm2e} 
\usepackage{tabularx}

\newtheorem{problem}{Problem}

\newcommand{\nop}[1]{}

\newcommand{\methodFont}{\textsf}

\newcommand{\actuated}{\methodFont{Actuated}\xspace}

\newcommand{\SOTL}{\methodFont{SOTL}\xspace}
\newcommand{\Formula}{\methodFont{Webster}\xspace}
\newcommand{\Greenwave}{\methodFont{GreenWave}\xspace}
\newcommand{\Maxband}{\methodFont{Maxband}\xspace}
\newcommand{\Maxpressure}{\methodFont{Max-pressure}\xspace}
\newcommand{\SCATS}{\methodFont{SCATS}\xspace}

\newcommand{\notationFont}{\mathsf}

\newcommand{\agent}{\notationFont{G}}

\newcommand{\action}{\notationFont{a}}

\newcommand{\staterep}{\notationFont{s}}

\newcommand{\reward}{\notationFont{r}}

\begin{document}
\title{A Survey on Traffic Signal Control Methods}

 \author{Hua Wei}
\affiliation{%
 \institution{College of Information Sciences and Technology, Penn State University}
\streetaddress{}
\city{University Park}
\state{PA}
 \postcode{16802}
\country{USA}}
\email{hzw77@ist.psu.edu}

 \author{Guanjie Zheng}
\affiliation{%
 \institution{College of Information Sciences and Technology, Penn State University}
\streetaddress{}
\city{University Park}
\state{PA}
 \postcode{16802}
\country{USA}}
\email{gjz5038@ist.psu.edu}

 \author{Vikash Gayah}
\affiliation{%
 \institution{Department of Civil Engineering, Penn State University}
\streetaddress{}
\city{University Park}
\state{PA}
 \postcode{16802}
\country{USA}}
\email{gayah@engr.psu.edu}

 \author{Zhenhui Li}
\affiliation{%
 \institution{College of Information Sciences and Technology, Penn State University}
\streetaddress{}
\city{University Park}
\state{PA}
 \postcode{16802}
\country{USA}}
\email{jessieli@ist.psu.edu}


\input{abstract}
\input{keyword}

\maketitle
\input{introduction}

\input{preliminary.tex}
\input{transportation}
\input{RL}

\input{conclusion}
\input{appendix}
\input{ack}

\bibliographystyle{ACM-Reference-Format}
\bibliography{bibliography}

\end{document}

%% file: abstract.tex

\begin{abstract}
Traffic signal control is an important and challenging real-world problem, which aims to minimize the travel time of vehicles by coordinating their movements at the road intersections. Current traffic signal control systems in use still rely heavily on oversimplified information and rule-based methods, although we now have richer data, more computing power and advanced methods to drive the development of intelligent transportation. With the growing interest in intelligent transportation using machine learning methods like reinforcement learning, this survey covers the widely acknowledged transportation approaches and a comprehensive list of recent literature on reinforcement for traffic signal control. We hope this survey can foster interdisciplinary research on this important topic. 

\end{abstract}

%% file: keyword.tex

%
%

%
%

\keywords{traffic signal control, transportation, reinforcement learning, deep learning, mobility data}

%% file: introduction.tex

\section{Introduction}
\label{sec:introduction}
Traffic congestion is a growing problem that continues to plague urban areas with negative outcomes to both the traveling public and society as a whole. These negative outcomes will only grow over time as more people flock to urban areas. In 2014, traffic congestion costs Americans over \$160 billion in lost productivity and wasted over 3.1 billion gallons of fuel~\cite{Econ14}. Traffic congestion was also attributed to over 56 billion pounds of harmful CO2 emissions in 2011~\cite{schrank20152015}. 
In the European Union, the cost of traffic congestion was equivalent to 1\% of the entire GDP~\cite{schrank2012tti}.
Mitigating congestion would have significant economic, environmental and societal benefits.
Signalized intersections are one of the most prevalent bottleneck types in urban environments, and thus traffic signal control plays a vital role in urban traffic management.

\begin{figure}[t]
\centering
\includegraphics[width=0.99\textwidth]{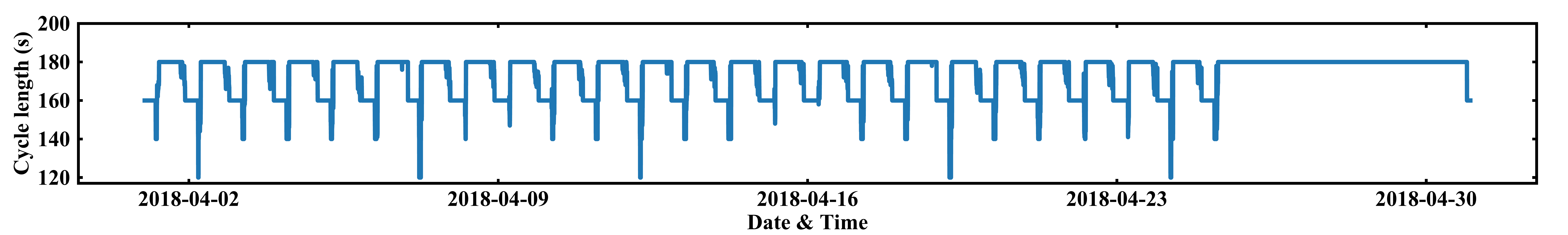} 
\caption{Traffic signal timing of a downtown intersection in Hangzhou, China. The x- and y-axis indicate the time and cycle length of the traffic signal. The cycle length rarely changes as time goes by.}
\label{fig:timing}
\end{figure}

\subsection{Current Situation}
In many modern cities today, the widely-used adaptive traffic signal control systems such as SCATS~\cite{SCATS} and SCOOT~\cite{hunt1981scoot,hunt1982scoot} heavily rely on manually designed traffic signal plans. Such manually set traffic signal plans are designed to be dynamically selected according to the traffic volume detected by loop sensors. However, many intersections do not have loop sensors installed or the loop sensors are poorly maintained. Moreover, the loop sensors are activated only when vehicles pass through them; thus, they can only provide partial information about the vehicle through them. As a result, the signal cannot perceive and react to real-time traffic patterns. Engineers need to manually change the traffic signal timings in the signal control system under certain traffic condition scenarios. Figure~\ref{fig:timing} shows the traffic signal timing at an intersection in a city of China, and the traffic signal timing rarely changes regardless of the real traffic changes throughout the day.

\subsection{Opportunities}
\emph{First, today we have much richer information that can be collected from various sources.} Traditional traffic signal control relies on data from loop sensors, which can only sense the vehicle passing. However, new data sources are quickly becoming available that can serve as input for traffic signal control purposes. For instance, street-facing surveillance cameras used for security purposes can also provide a more detailed depiction of  the traffic situation on nearby roads,  specifically on how many cars are waiting in the lane, how many cars are taking turns, where they are located, and how fast they are traveling. In addition, large-scale trajectory data can be collected from various sources such as navigation applications (e.g., Google Maps),  ride-sharing platforms (e.g., Uber), and GPS-equipped vehicles that share information with the nearby infrastructure (e.g., connected vehicles). Such data provide us with more insight into how vehicles arrive at intersections. We have reached a stage of sufficient mobility information that can describe the traffic dynamics in the city more clearly, which is an essential resource for us to improve the traffic control system.

\begin{figure}[htbp]
\centering
\includegraphics[width=0.7\columnwidth]{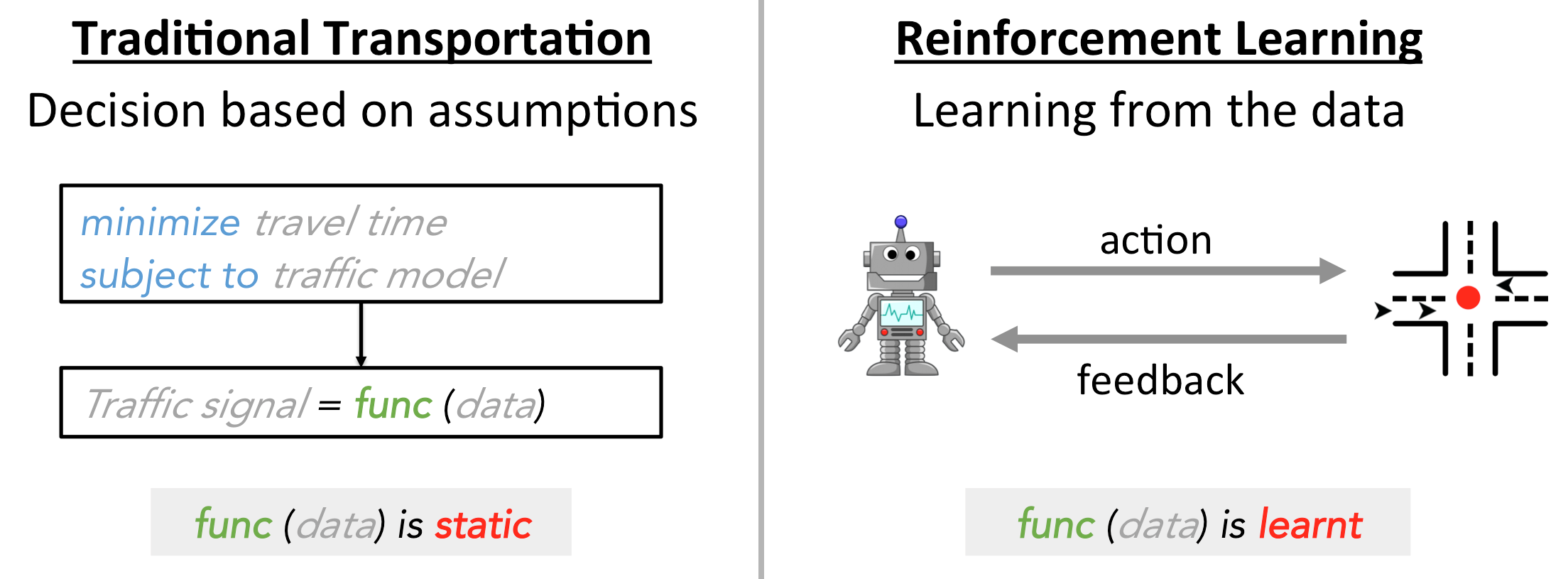}
\caption{Difference between traditional transportation approach and machine learning approach.}
\label{fig:optvsrl}
\end{figure}

\emph{Second, today we have the much stronger computing power and advanced computational models.} The typical approach that transportation researchers take is to cast traffic signal control as an optimization problem under certain assumptions about the traffic model, e.g., vehicles come in a uniform and constant rate~\cite{RPM04}.  Various (and sometimes strong) assumptions have to be made in order to make the optimization problem tractable. The key issue here is that these assumptions deviate from the real world and often do so significantly. As we know, real-world traffic condition evolves in a complicated way, affected by many factors such as driver's preference, interactions with vulnerable road users (e.g., pedestrians, cyclists, etc.), weather and road conditions. These factors can hardly be fully described in a traffic model.

On the other hand, machine learning techniques can directly learn from the observed data without making unrealistic assumptions about the model. However, typical supervised learning does not apply here because existing traffic signal control systems follow pre-defined signal plans so we do not have enough training data to differentiate good and bad traffic signal plan strategies. Instead, we have first to take actions to change the signal plans and then learn from the outcomes. This trial-and-error approach is also the core idea of reinforcement learning (RL). In essence, an RL system generates and executes different strategies (e.g., for traffic signal control) based on the current environment. It will then learn and adjust the strategy based on the feedback from the environment. This reveals the most significant difference between transportation approaches and our RL approaches, which is illustrated in Figure~\ref{fig:optvsrl}: in traditional transportation research, the model $func(data)$ is static; in reinforcement learning, the model is dynamically learned through trial-and-error in the real environment.

\subsection{Motivation of This Survey}

With the surge of AI technology and increasingly available city data, governments and industries are now actively seeking solutions to improve the transportation system. For example, in China, Alibaba and Didi Chuxing are working on using mobility data and advanced computing technology to enhance city transportation~\cite{didi, alibaba}. This survey could provide a useful reference for the industry when they revolutionize current traffic signal control systems by trying out the RL-based methods. Specifically, we discuss the learning approaches of recent RL-based methods with their weaknesses and strengths and benchmark the experiment settings of the existing works.

At the same time, with the recent success in reinforcement learning techniques, we see an increasing interest in academia to use reinforcement learning to improve traffic signal control~\cite{MaDH16}. However, most existing machine learning approaches tend to ignore classic transportation approaches and lack a good comparison with existing transportation approaches. This survey takes a comprehensive view of both machine learning and transportation engineering and hopes to facilitate this interdisciplinary research direction.

\subsection{Scope of This Survey}
In this survey, we will cover many classical or widely accepted transportation approaches in traffic signal control. We refer readers interested in comprehensive transportation approaches to~\cite{PDDK+03,LiWY14,RPM04}.
The application of reinforcement learning methods in traffic signal control is relatively new. Recent advances in deep reinforcement learning also arouses new applications of RL in traffic signal control problem. While~\cite{rlsurvey17} and~\cite{MaDH16} provide comprehensive surveys mainly on earlier studies before the popularity of deep reinforcement learning, in this survey, we will have comprehensive coverage in RL-based traffic signal control approaches, including the recent advances in deep RL-based traffic signal control methods.

%% file: preliminary.tex
\begin{figure}[htbp]
\centering
\includegraphics[width=0.9\columnwidth]{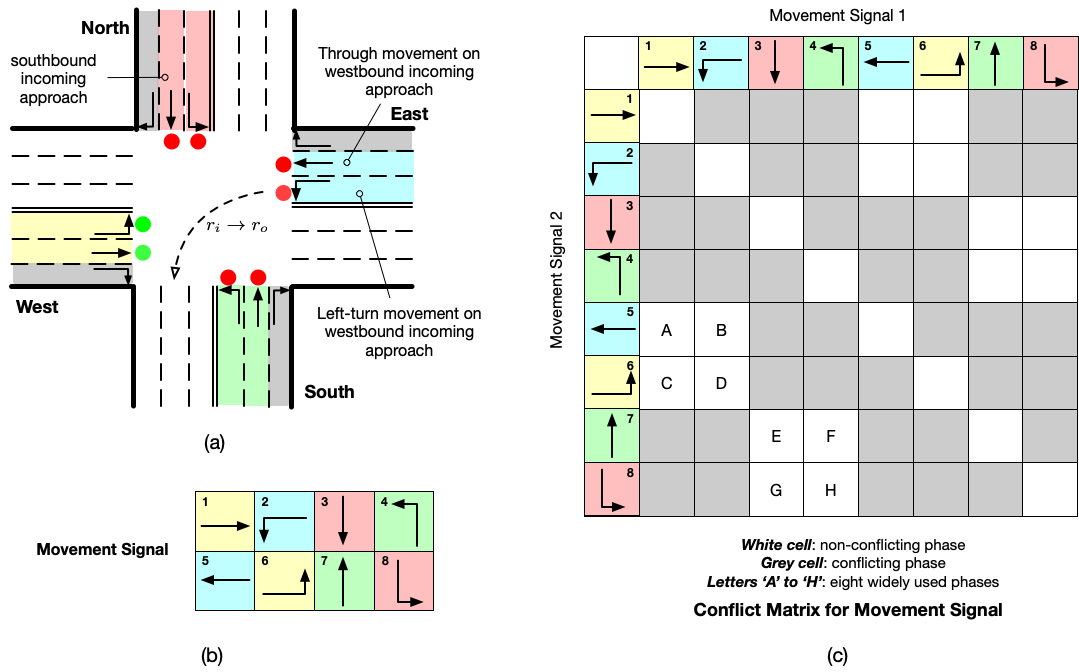}
\caption{Definitions of traffic movement and traffic signal phases.}
\label{fig:phase}
\end{figure}

\section{Preliminary}
\label{sec:basic-tlcs}
\subsection{Term Definition}

Terms on road structure and traffic movement:
\begin{itemize}
    \item \textbf{Approach}: A roadway meeting at an intersection is referred to as an approach. At any general intersection, there are two kinds of approaches:  incoming approaches and outgoing approaches. An incoming approach is one on which cars can enter the intersection; an outgoing approach is one on which cars can leave the intersection. Figure~\ref{fig:phase}(a) shows a typical intersection with four incoming approaches and outgoing approaches. The southbound incoming approach is denoted in this figure as the approach on the north side in which vehicles are traveling in the southbound direction.  
    \item \textbf{Lane}: An approach consists of a set of lanes. Similar to approach definition, there are two kinds of lanes:  incoming lanes and  outgoing lanes (also known as approaching/entering lane and receiving/exiting lane in some references~\cite{ElAb12,stevanovic2010adaptive}).
    \item \textbf{Traffic movement}: A traffic movement refers to  vehicles moving from an incoming approach to an outgoing approach, denoted as $(r_i\rightarrow r_o)$, where $r_i$ and $r_o$ is the incoming lane and the outgoing lane respectively. A traffic movement is generally categorized as left turn, through, and right turn. 

\end{itemize}

Terms on traffic signal:
\begin{itemize}
    \item \textbf{Movement signal}: A movement signal is defined on the traffic movement, with green signal indicating the corresponding movement is allowed and red signal indicating the movement is prohibited.
    For the four-leg intersection shown in Figure~\ref{fig:phase}(a), the right-turn traffic can pass regardless of the signal, and there are eight movement signals in use, as shown in Figure~\ref{fig:phase}(b).   
    \item \textbf{Phase}:  A phase is a combination of movement signals. 
    Figure~\ref{fig:phase}(c) shows the conflict matrix of the combination of two movement signals in the example in Figure~\ref{fig:phase}(a) and Figure~\ref{fig:phase}(b). The grey cell indicates the corresponding two movements conflict with each other, i.e. they cannot be set to `green' at the same time (e.g., signals \#1 and \#2). The white cell indicates the non-conflicting movement signals. All the non-conflicting signals will generate eight valid paired-signal phases (letters `A' to 'H' in Figure~\ref{fig:phase}(c)) and eight single-signal phases (the diagonal cells in conflict matrix). Here we letter the paired-signal phases only because in an isolated intersection, it is always more efficient to use paired-signal phases. When considering multiple intersections, single-signal phase might be necessary because of the potential spill back.
    \item \textbf{Phase sequence}:  A phase sequence is a sequence of phases which defines a set of phases and their order of changes.
    \item \textbf{Signal plan}: A signal plan for a single intersection is a sequence of phases and their corresponding starting time. Here we denote a signal plan as $(p_1,t_1)(p_2,t_2)\dots(p_i,t_i)\dots$, where $p_i$ and $t_i$ stand for a phase and its starting time.  
    \item \textbf{Cycle-based signal plan}: A cycle-based signal plan is a kind of signal plan where the sequence of phases operates in a cyclic order, which can be denoted as $(p_1,t^1_1)(p_2,t^1_2)\dots(p_N,t^1_N)(p_1,t^2_1)(p_2,t^2_2)\dots(p_N,t^2_N)\dots$, where  $p_1,p_2,\dots,p_N$ is the repeated phase sequence and $t^j_i$ is the starting time of phase $p_i$ in the $j$-th cycle. Specifically, $C^j=t^{j+1}_1-t^{j}_1$ is the cycle length of the $j$-th phase cycle, and $\{\frac{t^j_2-t^j_1}{C^j},\dots, \frac{t^j_N-t^j_{N-1}}{C^j}\}$ is the phase split of the $j$-th phase cycle. Existing traffic signal control methods usually repeats similar phase sequence throughout the day. 
\end{itemize}

\subsection{Objective}
The objective of traffic signal control is to facilitate safe and efficient movement of vehicles at the intersection. Safety is achieved by separating conflicting movements in time and is not considered in more detail here.  Various measures have been proposed to quantify efficiency of the intersection from different perspectives: 
\begin{itemize}
    \item Travel time. In traffic signal control, travel time of a vehicle is defined as the time different between the time one car enters the system and the time it leaves the system. One of the most common goals is to minimize the average travel time of vehicles in the network. 
    \item Queue length. The queue length of the road network is the number of queuing vehicles in the road network. 
    \item Number of stops. The number of stops of a vehicle is the total times that a vehicle experienced. 
    \item Throughput. The throughput is the number of vehicles that have completed their trip in the road network during a period. 
\end{itemize}

\subsection{Special Considerations}
In practice, additional attention should be paid to the following aspects:
 
\begin{enumerate}
\item Yellow and all-red time. A yellow signal is usually set as a transition from a green signal to a red one. Following the yellow, there is an all-red period during which all the signals in an intersection are set to red. The yellow and all-red time, which can last from 3 to 6 seconds, allow vehicles to stop safely or pass the intersection before vehicles in conflicting traffic movements are given a green signal. 

\item Minimum green time. Usually, a minimum green signal time is required to ensure pedestrians moving during a particular phase can safely pass through the intersection.

\item Left turn phase. Usually, a left-turn phase is added when the left-turn volume is above certain threshold.

\end{enumerate}

\nop{
\begin{figure}[t]
  \centering
  \begin{tabular}{cc}
   \includegraphics[width=0.42\textwidth]{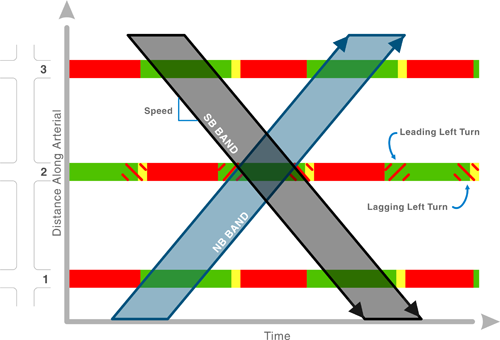} & 
   \includegraphics[width=0.48\textwidth]{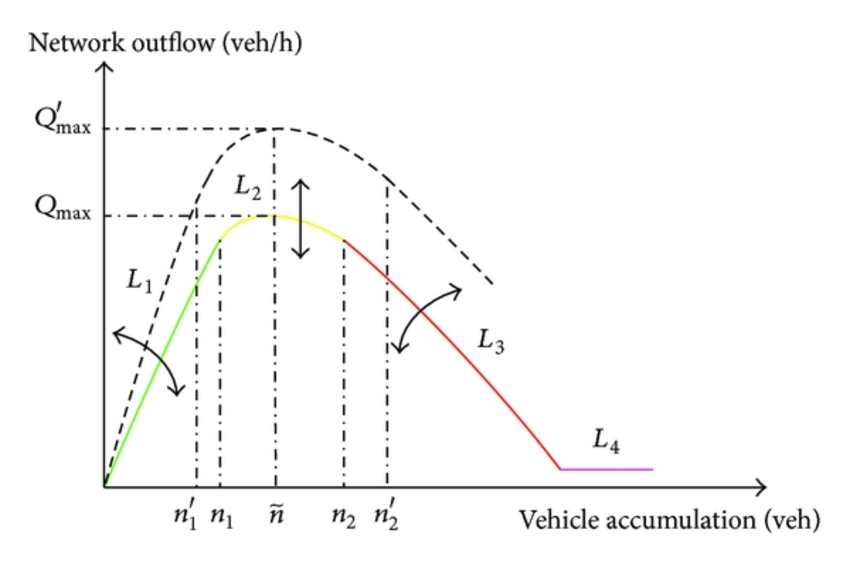}
   \end{tabular}
   \vspace{-1mm}
     \caption{Space-time diagram and Macroscopic fundamental diagram.}
\label{fig:diagram}
\end{figure}
}

%% file: transportation.tex
\section{Methods in transportation engineering}
\label{sec:transportation}

In this section, we introduce several selected classical methods in transportation field, which should be taken into comparison as baselines for RL-based methods. An overview of covered methods is shown in Table~\ref{table:overview}.
For a more comprehensive survey for the methods in transportation, we refer the interested readers to~\cite{RPM04,simulator11}.

\begin{table}[thb]
\caption{Overview of classic optimization-based transportation methods}
\begin{tabular}{ p{3cm} | p{4cm} | p{3cm} | p{4cm}}
\toprule
Method        & Prior Knowledge      & Data Input      & Output         \\ \midrule
\Formula       & phase sequence in a cycle   & traffic volume  & cycle-based signal plan for an individual intersection       \\  \midrule
\Greenwave,  \Maxband   & cycle-based signal plan at individual intersections  & traffic volume, speed limit, lane length  & offset for a cycle-based signal plan \\ \hline
\midrule
\actuated, \SOTL & phase sequence, phase change rule &  traffic volume & change to next phase according to the rule and data \\
\hline
\midrule
\Maxpressure    &  none & queue length   &  signal plan for all intersections  \\  \hline
\midrule
\SCATS   &  signal plan for all intersections  &  traffic volume  & adjusted signal plans  \\ 
\bottomrule
\end{tabular}
\label{table:overview}
\end{table}

 \subsection{\Formula (single intersection)}
 For a single (isolated) intersection, the traffic signal plan in transportation engineering usually consists of a pre-timed cycle length , fixed cycle-based phase sequence, and phase split. \Formula method~\cite{koonce2008traffic} is one of the widely-used method in field to calculate the cycle length and phase split for a single intersection. Assuming the traffic flow is uniform during a certain period (i.e., past 5 or 10 minutes), it has a closed-form solution shown in Eq.~\eqref{eq:webster-cycle} and~\eqref{eq:webster-split} to generate the optimal cycle length and phase split for a single intersection that minimizes the travel time of all vehicles passing the intersection. 
 
 The calculation of the desired cycle length $C_{des}$ relies on Webster's Equation:
 \begin{equation}
\label{eq:webster-cycle}
    C_{des}(V_c) = \frac{N\times t_L}{1-\frac{V_c}{\frac{3600}{h} \times PHF \times (v/c)}}
\end{equation}
where $N$ is the number of phases; $t_L$ is the total loss time per phase, which can be treated as a parameter related to the all-red time and the acceleration and deceleration of vehicles; parameter $h$ is saturation headway time (seconds/vehicle), which is the smallest time interval between successive vehicles passing a point; $PHF$ is short for peak hour factor, which is a parameter measuring traffic demand fluctuations within the peak hour; and the parameter $v/c$ is desired volume-to-capacity ratio, which indicates how busy the intersection is in a signal timing context. These parameters usually vary in different traffic conditions and are usually selected based on empirical observations and agency standards. The equation represents a function of $V_c$, where $V_c=\Sigma^N_i V^i_c$ is the sum of all critical lane volumes, which indicates how busy the intersection is in the signal timing context, and $V^i_c$ stands for the critical lane volumes for phase $i$, where a critical lane is the approaching lane with the highest ratio of traffic flow to saturation flow in a phase, usually indicated by the queue length.

Once the cycle length is decided, the green split (i.e., the green time over the cycle length) is then calculated to be proportional to the ratios of critical lane volumes served by each phase, as indicated in Eq.~\eqref{eq:webster-split}:
 \begin{equation}
\label{eq:webster-split}
   \frac{t_i}{t_j}=\frac{V^i_c}{V^j_c}
\end{equation}
where $t_i$ and $t_j$ stands for the phase duration for phase $i$ and $j$. 

Eq.~\eqref{eq:webster-cycle} and~\eqref{eq:webster-split} are typically applied using aggregated data to develop fixed-time plans. However, these equations can also be modified for real-time application. For example, when the traffic is uniform, the \Formula method can be proved to minimize the travel time of all vehicles passing the intersection or maximize the intersection capacity. By collecting data for a short time period and assuming no fluctuation in traffic demand and no redundancy left in a cycle, i.e., setting both $PHF$ and $v/c$ to 1, Eq.~\eqref{eq:webster-split} can be re-organized as 
\begin{equation}
(1-\frac{N\times t_L}{C_{des}})\frac{3600}{h} = V_c.
\label{eq:transportation-theory-simplified}
\end{equation}

The left side of Eq.~\eqref{eq:transportation-theory-simplified} is the proportion of time utilized for traffic movements at the intersection multiplied by the saturation flow of vehicles, which is the capacity of the intersection. When capacity equals to the traffic demand $V_c$, cycle length $C_{des}$ is the minimum value that tightly satisfies the traffic demands. 
For a given traffic demand, if a signal cycle length smaller than $C_{des}$ is applied, the queue length will keep increasing and the intersection gets over-saturated. If a signal cycle length larger than $C_{des}$ is applied, the mean delay for each vehicle at the intersection will grow linearly with the cycle length.

 \subsection{\Greenwave} 

While \Formula method generates the signal plan for a single intersection, the offsets (i.e., starting time between phase cycles at adjacent intersections) between the signal timings for adjacent traffic signals should also be optimized as signals are often in close proximity. Failure to do so can lead to decisions being made at one signal that can deteriorate traffic operations at another. 
 
\Greenwave~\cite{RPM04} is the most classical method in transportation field to implement coordination, which aims to optimize the offsets to reduce the number of stops for vehicles traveling along one certain direction. Given the signal plan (i.e., cycle length and phase split) of individual intersections, \Greenwave requires all intersections to share the same cycle length, which is the maximum value of the given cycle lengths for individual intersections. The offsets between intersections are calculated by the following equation:
 \begin{equation}
\label{eq:offset}
   \Delta t_{i,j}=\frac{L_{i,j}}{v}
\end{equation}
 where $L_{i,j}$ is the road length between intersection $i$ and $j$, and $v$ is the expected travel speed of vehicles on the road. 
 
 This method can form a green wave along the designated direction of traffic where vehicles traveling along that direction can benefit from a progressive cascade of green signals without stopping at any intersections.  However, \Greenwave only optimizes for unidirectional traffic. 
 
 \subsection{\Maxband}
Another typical approach, \Maxband~\cite{little1981maxband}, also takes the signal plans for individual intersections as input and optimizes the offsets for adjacent traffic signals. Different from \Greenwave, it aims to reduce the number of stops for vehicles traveling along \textit{two} opposite directions through finding a maximal bandwidth based on the signal plans of individual intersections within the system.  
The bandwidth for one direction is a portion of time that the synchronized green wave lasts in a cycle length. A larger bandwidth implies that more traffic along one direction can progress through the signals without stops. 

Like \Greenwave, \Maxband~\cite{little1981maxband} requires all intersections to share the same cycle length, which equals to the maximum value of all cycle lengths of intersections. Then \Maxband formulates a mixed integer linear programming model to generate a symmetric, uniform-width bandwidth, subject to the following physical constraints:

\begin{figure}
    \centering
    \includegraphics[width=0.9\textwidth]{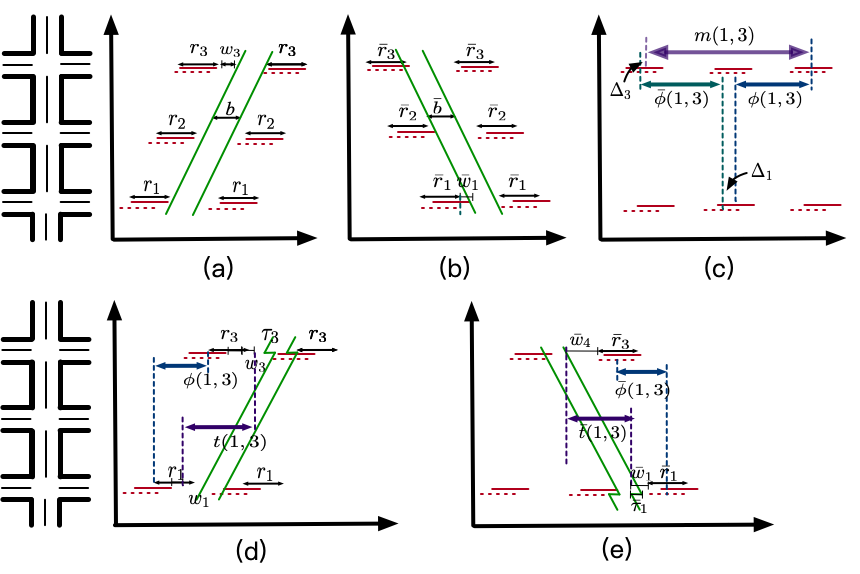}
    \caption{Illustration of the constraints in \Maxband under a three-intersection arterial network. Red solid/dotted lines indicate the red signal for inbound/outbound direction. The green band indicates the green wave. (a) - (b): Bandwidth constraints of each individual intersection on inbound and outbound direction. (c): Temporal constraint between two intersections. (d) - (e): Spatial constraints between two intersections.}
    \label{fig:maxband}
\end{figure}

\begin{itemize}
    \item Constraints on the bandwidth of individual intersections. For simplicity, all the time intervals in the following refers to the ratio of time over the cycle length $C$.
    \begin{itemize}
    \item For each direction (here we use inbound and outbound to differentiate two designated directions), the green time should be greater than the bandwidth. We have:
        \begin{subequations}
        \label{eq:MB-1}
        \begin{align}
            \label{eq:MB-1-1}
             w_i+b\leq 1-r_i,\ w_i>0 \\
             \label{eq:MB-1-2}
             \bar{w}_i+\bar{b}\leq 1-\bar{r}_i,\ \bar{w}_i>0
        \end{align}
        \end{subequations}
        Here, $w_i/\bar{w}_i$ denotes the time interval between the end of red time and the start of bandwidth on inbound/outbound direction, $b$ is the bandwidth variable, and $r_i/\bar{r}_i$ is the red time on inbound/outbound direction.
    \item For two different directions (inbound and outbound), usually the bandwidths for inbound and outbound are set equal:
        \begin{equation}
                \label{eq:MB-bandwidth}
                    b = \bar{b}
        \end{equation}
    \end{itemize}
    \item Temporal and spatial constraints on offsets between two intersections $i$ and $j$. 
        \begin{itemize}
            \item Temporal constraints. 
                \begin{equation}
                \label{eq:MB-2}
                    \phi(i,j) +\bar{\phi}(i,j) + \Delta_i - \Delta_j = m(i,j)
                \end{equation}
                Here, $\phi/\bar{\phi}$ is the inbound/outbound offset between intersections, $\Delta$ is the intra-intersection offset between the inbound and outbound red time, $m(i,j)$ is an integer variable, indicating a multiple of cycle length.
            \item Spatial constraints. The travel time of a vehicle starting from one intersection to another satisfies a function of the offset between them, the queue clearance time at the destined intersection and their red time.
                \begin{subequations}
                \begin{align}
                \label{eq:MB-3-1}
                    \phi(i,j)+0.5*r_j+w_j+\tau_j = 0.5*r_i + w_i+t(i,j) \\
                     \bar{\phi}(i,j)+0.5*\bar{r}_j+\bar{w}_j = 0.5*\bar{r}_i + \bar{w}_i-\bar{\tau}_i+\bar{t}(i,j)
                     \label{eq:MB-3-2}
                \end{align}
                \end{subequations}
            Here, $t/\bar{t}$ is the travel time inbound/outbound between intersections, which is relevant to the road length and traffic speed. $\tau/\bar{\tau}$ is the queue clearance time at an intersection, which is relevant to the turn-in and generated traffic.
            \end{itemize}
\end{itemize}

In summary, \Maxband aims to find $b$, $\bar{b}$ $w_i$, $\bar{w}_i$, $m_i$ that:
\begin{equation}
    \begin{split}
  maximize \ \ &  b  \\
  subject\ to\ \ &Eq.~\eqref{eq:MB-1-1},~\eqref{eq:MB-1-2},~\eqref{eq:MB-bandwidth},~\eqref{eq:MB-2},~\eqref{eq:MB-3-1},~\eqref{eq:MB-3-2} \\
    &m_i \in \mathbb{N}\\
    &b,\ \bar{b},\ w_i,\ \bar{w}_i \geq 0, i=1,\dots,n
    \end{split}
\end{equation}

A number of significant extensions of \Maxband have been introduced based on the original method~\cite{little1981maxband} in order to consider a variety of new aspects. ~\cite{gartner1991multi} extends to include asymmetric bandwidths in the opposing direction, variable left-turn phase sequence, as well as decisions on cycle time length and link specific progression speeds. ~\cite{stamatiadis1996multiband} presents the new multi-band, multi-weight approach, which also incorporates all previous decision capabilities.

\subsection{Actuated Control}
Actuated control measures the ``requests'' for a green signal from the current phase and other competing phases, then based on some rules to decide whether to keep or change the current phase. 
The definitions of ``request'' for the current phase and other competing phases are:
\begin{table}[htb]
\caption{Rules for actuated control}
\label{tab:rule-actuated}
\begin{tabular}{p{2.5cm}|p{2.5cm} |p{9cm}}
\toprule
\begin{tabular}[c]{@{}l@{}}Request from\\ current phase?\end{tabular} & \begin{tabular}[c]{@{}l@{}}Request from\\ other phases?\end{tabular} & Action   \\ \midrule
\multirow{4}{*}{Yes}     & \multirow{2}{*}{Yes}  & Fully-actuated control:If the duration for the current phase is larger than a threshold, change to the next phase; otherwise, keep the current phase \\ 
                          &                        & Semi-actuated control: Keep the current phase  \\ \cline{2-3}
                        & No                  & Keep the current phase                 \\ \midrule
\multirow{4}{*}{No}     & Yes                   & 
\begin{tabular}[c]{@{}l@{}}
Change to the next phase  
\end{tabular}  \\ \cline{2-3}
                       & \multirow{2}{*}{No} & Fully-actuated: Keep the current phase \\  
                          &                        & Semi-actuated: Change to the default phase (usually set as green signal for the main road   \\  
\bottomrule                                                
\end{tabular}
\end{table}

\begin{itemize}
    \item The request on the current phase to extend the green signal time is generated when the duration for the current phase does not reach a minimum time period, or there is a vehicle on the incoming lane of the current phase and is within close distance to the intersection. And we call this vehicle is ``approaching the green signal'' in short.
    \item The request on the competing phases for a green signal is generated when the number of waiting vehicles in the competing phases is larger than a threshold.
\end{itemize}

Based on the difference in rules, there are two kinds of actuated control: fully-actuated control and semi-actuated control. The rules for the decision of keep or changing the current phase are listed in Table~\ref{tab:rule-actuated}.

\subsection{Self-organizing Traffic Light Control}

\begin{table}[htb]
\caption{Rules for \SOTL control}
\label{tab:sotl}
\begin{tabular}{l|l|l}

\toprule
\begin{tabular}[c]{@{}l@{}} Request from\\ current phase?\end{tabular} & \begin{tabular}[c]{@{}l@{}} Request from\\ competing phases?\end{tabular} & Action                                  \\ \midrule
\multirow{2}{*}{Yes}      & Yes      & \multirow{2}{*}{Keep the current phase} \\ \cline{2-2}
                           & No    &                                         \\ \midrule
\multirow{2}{*}{No}     & Yes     & Change to the next phase                \\ \cline{2-3} 
                           & No    & Keep the current phase                  \\ \bottomrule
\end{tabular}
\end{table}

Self-Organizing Traffic Light Control (\SOTL) is basically the same as fully-actuated control with additional demand responsive rules~\cite{cools2013self, Carl04}. The main difference between \SOTL and fully-actuated control is on the definition of request on the current phase (although they both require a minimum green phase duration): in fully-actuated control, the request on the current phase will be generated whenever there is a vehicle approaching the green signal, while in \SOTL, the request will not be generated unless the number of vehicles approaching the green signal is larger than a threshold which is not necessarily one. Specifically, the rules for \SOTL control are shown in Table~\ref{tab:sotl}.

\subsection{\Maxpressure}

\begin{figure}[htb]
    \centering
    \includegraphics[width=0.6\textwidth]{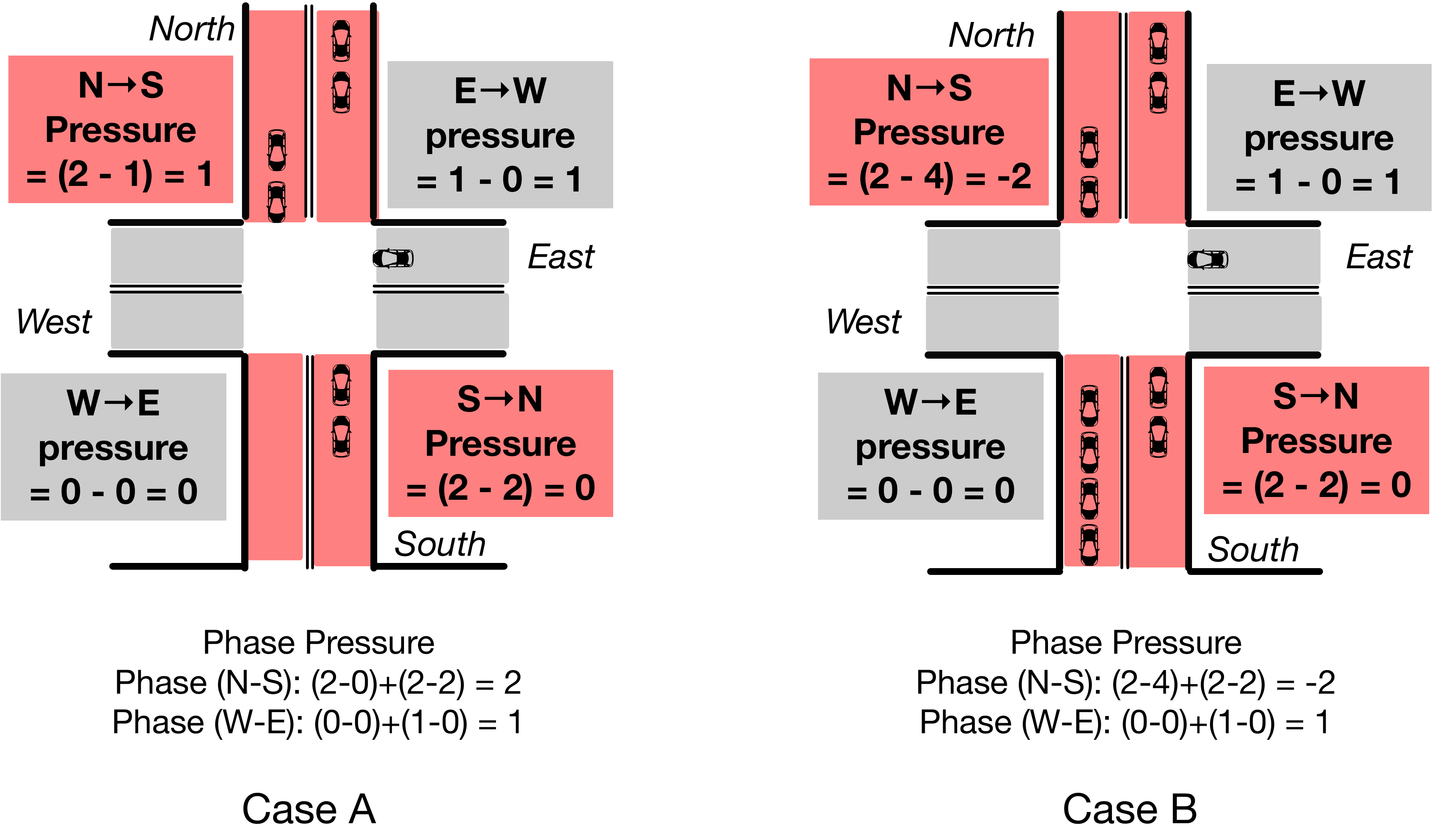}
    \caption{Illustration of max pressure control in two cases. In both cases, there are four movement signals: North$\rightarrow$South, South$\rightarrow$North, East$\rightarrow$West and West$\rightarrow$East and there are two phases: $Phase(N-S)$ which sets green signal in the North$\rightarrow$South and South$\rightarrow$North direction, and $Phase(N-S)$ which sets green signal in the East$\rightarrow$West and West$\rightarrow$East direction. In Case A, \Maxpressure selects $Phase(N-S)$ since the pressure of $Phase(N-S)$ is higher than $Phase(W-E)$; in Case B, \Maxpressure selects $Phase (W-E)$.}
    \label{fig:pressure}
\end{figure}

\Maxpressure control~\cite{varaiya2013max} aims to reduce the risk of over-saturation by balancing queue length between neighboring intersections by minimizing the ``pressure'' of the phases for an intersection. The concept of pressure is illustrated in Figure~\ref{fig:pressure}. Formally, the pressure of a \textit{movement signal} can be defined as the number of vehicles on incoming lanes (of the traffic movement) minus the number of vehicles on the corresponding outgoing lanes; the pressure of a \textit{phase} is defined as the difference between the total queue length on incoming approaches and outgoing approaches.  By setting the objective as minimizing the pressure of phases for individual intersections, \Maxpressure is proved to maximize the throughput of the whole road network. For readers interested in the proof, please refer to~\cite{varaiya2013max}. 

\Maxpressure control proposed in~\cite{varaiya2013max} is formally summarized in Algorithm~\ref{alg:mp}. From line 3 to line 7, this method selects the phase with the maximum pressure, activates it as the next phase and keeps the selected phase for a given period of time $t_{min}$. 

\begin{algorithm}[H]
\KwIn{Current phase time $t$, minimum phase duration length $t_{min}$}
\SetAlgoLined
\ForAll{timestamp}
{
    $t = t + 1$\\
    \If{$t>=t_{min}$}
    {
        Calculate the pressure $P_i$ for each phase $i$; \\
        Set the next phase as $\arg\max_{i} \{ P_i \}$ ; \\ 
        $t=0$\;
    }
}

 \caption{\Maxpressure Control}
 \label{alg:mp}
\end{algorithm}

\subsection{SCATS}
SCATS~\cite{lowrie1990scats} (Sydney Coordinated Adaptive Traffic System) takes pre-defined signal plans (i.e., cycle length, phase split and offsets) as input and iteratively selects from these traffic signals according to the pre-defined performance measurements. Its measurement like the degree of saturation ($DS$) is detailed as follows:

\begin{equation}
\label{eq:scats-ds}
    DS = \frac{g_{E}}{g}=\frac{g-(h'-h\times n)}{g}
\end{equation}

where $g$ is the available green signal time (in seconds), $g_{E}$ is the effective green time during which there are vehicles passing through the intersection and equals to available green signal time minus the wasted green time; and the wasted green time is calculated through the detection - $h'$ is the detected total time gap, $n$ is the detected number of vehicles, and  $h$ is the unit saturation headway (seconds) between vehicles, which stands for the smallest time interval between successive vehicles passing a point and usually set with expert knowledge.

The phase split (i.e., the phase time ratio of all phases), cycle length and offsets are selected from several pre-defined plans using similar mechanisms. Take the selection mechanism of the phase split as an example. As shown in Algorithm~\ref{alg:SCATS}, the algorithm first calculates $DS$ for the current signal plan using Eq.~\eqref{eq:scats-ds} at the end of each cycle (line 2 to 4). Then from line 6 to line 10, the algorithm infers the $DS$ of other signal plans which are not applied in the current cycle using the following equation:
\begin{equation}
\label{eq:scats-select}
\bar{DS}^j_{p} = \frac{DS_{p}\times g_{p}}{g^j_{p}}
\end{equation}
where $DS_{p}$ and $g_{p}$ are the the degree of saturation and green signal time for phase $p$ in the current signal plan, and $g^j_{p}$ is the degree of saturation for phase $p$ in signal plan $j$. Then the algorithm selects the signal plan with the optimal $DS$.

\begin{algorithm}
\caption{Selection of Phase Split in SCATS}
\label{alg:SCATS}
\KwIn{Candidate signal plan set $A=\{a_j| j=1,\dots,N\}$ and current plan $a_c$}
\KwOut{Signal plan for next cycle}
(Calculate $DS_{p}$ for each phase $p\in\{1,\dots,P\}$ in the current signal plan $a_c$)\\
\ForAll{$p$ in $\{1,\dots,P\}$}
{
Calculate $DS_{p}$ using Eq.~\eqref{eq:scats-ds} \;
}\
(Infer $\bar{DS}^j_{p}$ for all candidate signal plans)\\
\ForAll {$a_j$ in $A$}
{
    \ForAll{$p$ in $\{1,\dots,P\}$}
    {
       Calculate $\bar{DS}^j_{p}$ using Eq.~\eqref{eq:scats-select} \;
    }
}\
Return the signal plan $a_j$ that minimizes $\Sigma^P_i \bar{DS}^j_{i}$;
\end{algorithm}

%% file: RL.tex

\section{Reinforcement Learning based Traffic Signal Control}
\label{sec:rl-tlcs}
Recently, different artificial intelligent techniques have been proposed to control the traffic signal, like fuzzy logic algorithms~\cite{GoSr10}, swarm intelligence~\cite{Teod08}, and reinforcement learning~\cite{Wier00,MaDH16,KWBV08,ElAb12}. Among these technologies, RL is more trending these years.

\subsection{Preliminaries}

\subsubsection{Basic concepts}

In order to solve the traffic signal control problem using RL, we first introduce the formulation of RL in Problem~\ref{prob:RLI}, then we introduce how traffic signal control fits the RL setting. We will also introduce the formulation of multi-agent RL (MARL) in the context of traffic signal control.

\begin{problem}[RL Framework] 
\label{prob:RLI}
Usually a single-agent RL problem is modeled as a Markov Decision Process (MDP) $<\mathcal{S}, \mathcal{A}, P, R, \gamma>$, where $\mathcal{S}, \mathcal{A}, P, \mathcal{R}, \gamma$ are the set of state representations, the set of action, the probabilistic state transition function, the reward function, and the discount factor respectively. Given two sets $\mathcal{X}$ and $\mathcal{Y}$, we use $\mathcal{X}\times \mathcal{Y}$ to denote the Cartesian product of $\mathcal{X}$ and $\mathcal{Y}$, i.e., $\mathcal{X}\times \mathcal{Y}=\{ (x,y ) | x\in \mathcal{X},  y\in\mathcal{Y}\}$.
The definitions are given as follows:
\begin{itemize}
    \item $\mathcal{S}$: At time step $t$, the agent observes state $\staterep^t\in\mathcal{S}$.
    \item $\mathcal{A}, P$: At time step $t$, the agent takes an action $\action^t\in\mathcal{A}$, which induces a transition in the environment according to the state transition function
    \begin{equation}
        \label{eq:transition}
        P(\staterep^{t+1}|\staterep^t,\action^t): \mathcal{S}\times\mathcal{A} \rightarrow \mathcal{S}
    \end{equation}
    \item $R$: At time step $t$, the agent obtains a reward $r^t$ by a reward function.
        \begin{equation}
            R(\staterep^t,\action^t): \mathcal{S}\times\mathcal{A} \rightarrow \mathbb{R}
        \end{equation}
    \item $\gamma$: The goal of an agent is to find a policy that maximizes the expected return, which is the discounted sum of rewards:
        \begin{equation}
        \label{eq:expeted-return}
            G^t := \sum_{i=0}^\infty\gamma^i \reward^{t+i}
        \end{equation}
    where the discount factor $\gamma \in [0, 1]$ controls the importance of immediate rewards versus future rewards. Here, we only consider continuing agent-environment intersections which do not end with terminal states but goes on continually without limit.
\end{itemize}
\end{problem}

\begin{definition} [Optimal policy and optimal value functions]
Solving a reinforcement learning task means, roughly, finding an optimal policy $\pi^*$ that maximizes expected return. While the agent only receives reward about its immediate, one-step performance, one way to find the optimal policy $\pi^*$ is by following an optimal \textit{action-value function} or \textit{state-value function}.
The action-value function (Q-function) of a policy $\pi$, $Q^{\pi}: \mathcal{S}\times\mathcal{A} \rightarrow \mathbb{R}$, is the expected return of a state-action pair $Q^{\pi}(\staterep,\action) = \mathbb{E}_{\pi}[G^t| \staterep^t=\staterep, \action^t=\action]$.

The optimal Q-function is defined as $Q^*(\staterep,\action)=\max_{\pi}Q^{\pi}(\staterep,\action)$. It satisfies the Bellman optimality equation:
\begin{equation}
\label{eq:action-value}
   Q^{*}(\staterep^t,\action^t) = \sum_{\staterep^{t+1}\in \mathcal{S}} P(\staterep^{t+1}|\staterep^t,\action^t)[r^t+\gamma\max_{\action^{t+1}}Q^{*}(\staterep^{t+1},\action^{t+1})], \forall \staterep^t \in \mathcal{S}, \action^t \in \mathcal{A}
\end{equation}

The state-value function of a policy $\pi$, $V^{\pi}: \mathcal{S}\rightarrow \mathbb{R}$, is the expected return of a state $V^{\pi}(\staterep)=\mathbb{E}_{\pi}[G^t| \staterep^t=\staterep]$. The optimal state-value function is defined as $V^{*}(\staterep)=\max_{\pi}V^{\pi}(\staterep)$. It satisfies the Bellman optimality equation:
\begin{equation}
\label{eq:state-value}
    V^{*}(\staterep^t) = \max_{\action^t \in \mathcal{A}} \sum_{\staterep^{t+1}\in \mathcal{S}} P(\staterep^{t+1}|\staterep^t,\action^t)[\reward+\gamma v_*(\staterep^{t+1})], \forall \staterep^t \in \mathcal{S}
\end{equation}
\end{definition}

\begin{figure}[t]
  \centering
   \includegraphics[width=0.85\textwidth]{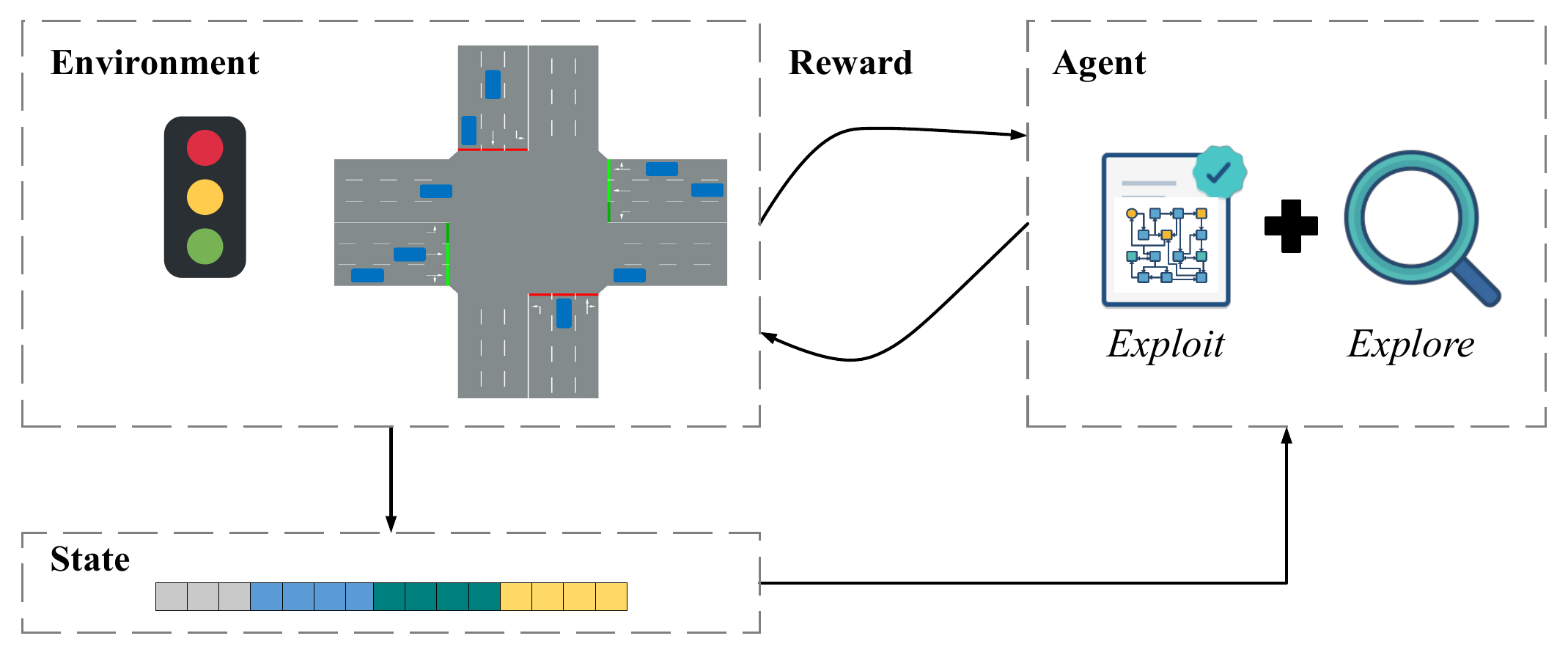}
   \vspace{-1mm}
     \caption{Reinforcement learning framework for traffic light control.}
     \vspace{-3mm}
    \label{fig:system}
\end{figure}

\begin{example} [Isolated Traffic Signal Control Problem]
Figure~\ref{fig:system} illustrates the basic idea of the RL framework in traffic light control problem. The environment is the traffic conditions on the roads, and the agent $\agent$ controls the traffic signal. At each time step $t$, a description of the environment (e.g., signal phase, waiting time of cars, queue length of cars, and positions of cars) will be generated as the state $\staterep_t$. The agent will make a prediction on the next action $\action^{t}$ to take that maximizes the expected return defined as Eq.~\eqref{eq:expeted-return},
where the action could be changing to a certain phase in a single intersection scenario. The action $\action^t$ will be executed in the environment, and a reward $\reward^{t}$ will be generated, where the reward could be defined on the traffic condition of the intersection. Usually, during the decision process, the policy that the agent takes combines the exploitation of learned policy and exploration of a new policy.
\end{example}

\begin{problem}[Multiagent RL Framework]
The generalization of the MDP to the multi-agent case is the stochastic game (SG). A stochastic game is defined by a tuple $\Gamma=<\pmb{S},\pmb{P},\pmb{A},\pmb{R},\pmb{O},\mathit{N},\gamma>$, where $\pmb{S},\pmb{P},\pmb{A},\pmb{R},\pmb{O},\mathit{N},\gamma$ are the sets of states, transition probability functions, joint actions, reward functions, private observations, number of agents and a discount factor respectively.  The definitions are given as follows:
\begin{itemize}
    \item $N$: $N$ agents identified by $i\in\pmb{I}=\{1,\dots N\}$.
    \item $\pmb{S},\pmb{O}$: At each time step $t$, agent $i$ draws observation $o^t_i\in \pmb{O}$ correlated with the true environment state $\staterep^t\in \pmb{S}$ according to the observation function $\pmb{S}\times \pmb{I}\rightarrow \pmb{O}$. 

    \item $\pmb{P},\pmb{A}$. Agent $i$'s action set $\pmb{A}_i$ is defined as a group of phases. At time step $t$, each agent takes an action $\action_i^t\in \pmb{A}_i$, forming a joint action $\pmb{a}^t=\action_1,\dots,\action_N$, which induces a transition in the environment according to the state transition function
    \begin{equation}
    \label{eq:transition-MARL}
    \pmb{P}(\staterep^{t+1}|\staterep^t,\pmb{a}^t): \pmb{S}\times \pmb{A}_1 \times \dots \times \pmb{A}_N \rightarrow \Omega(\pmb{S})
    \end{equation}
    where $\Omega(\pmb{S})$ denotes the space of state distributions.
    
    \item $\pmb{R}$: In a stochastic game setting, the reward an agent obtains is also influenced by the actions of other agents. Therefore, at time $t$, each agent $i$ obtains rewards $r_i^t$ by a reward function
    \begin{equation}
    R_i(\staterep^t,\pmb{a}^t): \pmb{S}\times \pmb{A}_1 \times \dots \times \pmb{A}_N \rightarrow \mathbb{R}
    \end{equation}

    \item $\gamma$: Intuitively, the joint actions have long-term effects on the environment. Each agent $i$ chooses an action following a certain policy $\pi_i$, aiming to maximize its total reward, $G^t := \sum_{j=0}^\infty\gamma^j \reward_i^{t+j}$, where the discount factor $\gamma \in [0, 1]$ controls the importance of immediate rewards versus future rewards.
\end{itemize}
\end{problem}

\begin{example} [Multi-intersection Traffic Signal Control Problem]
For a network with multiple intersections, the agents are defined as the signal controllers for $N$ intersections in the environment. The goal of traffic signal agents controlled with RL is to learn the optimal policies of each agent as well as to optimize the traffic condition of the whole environment. At each timestep $t$, each agent $i$ observes part of the environment as the observation $o_i^t$. The agents will make predictions on the next actions $\pmb{a}^{t}$ to take. In the real world, $\pmb{A}_i$ is mostly pre-defined, i.e., the traffic signal can only change among several phases. The actions will be executed in the environment, and the reward $\reward_i^{t}$ will be generated, where the reward could be defined on the level of individual intersections or a group of intersections within the environment. 
\end{example}

\subsubsection{Basic components of RL for traffic signal control}
There are four main components to formulate the problem under the framework of RL:

\begin{itemize}
\item Reward design. As RL is learning to maximize a numerical reward, the choice of reward determines the direction of learning.   
\item State design. State captures the situation on the road and converts it to values. Thus the choice of states should sufficiently describe the environment.
\item Selection of action scheme. Different action schemes also have influences on the performance of traffic signal control strategies. For example, if the action of an agent is defined as ``which phase to change to'', the traffic signal will have a more flexible phase sequence than defining the action as ``keep current phase or change to the next phase in a cycle''.  
\item Coordination strategy. How to achieve coordination is one of the challenges that complicate the signal control problem. In urban environments, signals are often in close proximity, and vehicles departing from one signal influence the arrival pattern of vehicles to the next downstream intersection. Thus, optimizing of signal timings for adjacent traffic signals must be done jointly. 
\end{itemize}

\subsection{RL Formulation}
A key question for RL is how to formulate the RL setting, i.e., the reward, state and action definition. In recent studies~\cite{Wier00,VaOl16,wei2018intellilight}, a typical reward definition for traffic signal control is a weighted linear combination of several components such as queue length, waiting time and delay. The state features include components such as queue length, number of cars, waiting time, and current traffic signal phase. In recent work~\cite{VaOl16,wei2018intellilight}, images of vehicles' positions on the roads are also considered in the state and fed into deep neural networks to learn the control policies. For more discussions on the reward, state, and actions, we refer interested readers to~\cite{rlsurvey17,MaDH16, el2011comprehensive}.

\subsubsection{State Definitions}
\label{sec:state}

\begin{table}[htbp]
\caption{Elements in state definitions}
\begin{center}
\begin{tabular}{p{3cm} p{12cm}}
\toprule
Element  & References \\
\midrule
Queue length & \cite{abdoos2011traffic,AMB14,BGS10,CBYT11,ElAb10,ElAb12,EAA13,MaDH16,SCGC08,wei2018intellilight,XXL13,BPT14,PBTB+13,PraB11,aslani2017adaptive,aslani2018traffic,GCN18,chu2019multi,lit2019,frep2019,colight2019, demolight, metaLight, mpLight} \\ \midrule
Waiting time & \cite{wei2018intellilight,chu2019multi} \\ \midrule
Volume & \cite{SaCa10,BGS10,ElAb10,wei2018intellilight,Casa17,aslani2017adaptive,aslani2018traffic,presslight2019}\\ \midrule
Delay &  \cite{ALUK10}\\ \midrule
Speed & \cite{ElAb10,Casa17,GCN18}  \\  \midrule
Phase duration & \cite{EAA13,MaDH16,BPT14,PBTB+13,PraB11} \\ \midrule
Congestion & \cite{BWKG10,IKKK06,SSPN+05} \\ \midrule
Position of vehicles & \cite{wei2018intellilight,Wier00,BWKG10,IKKK06,KWBV08,SSPN+05,WVVK04-2,WVVK04,VaOl16,KhGo12,KhGE12,mousavi2017traffic} \\ \midrule

Phase  & \cite{EAA13,MaDH16,SCGC08,wei2018intellilight,aslani2017adaptive,aslani2018traffic,presslight2019,colight2019,lit2019,frep2019,demolight,metaLight,mpLight} \\ 
\bottomrule
\end{tabular}
\label{tab:state-definitions}
\end{center}
\end{table}

At each time step, the agent receives some quantitative descriptions of the environment as the state representation.
As is shown in Table~\ref{tab:state-definitions}, various kinds of elements have been proposed to describe the environment state in the traffic signal control problem:
\begin{itemize}
    \item Queue length. Queue length of a lane is the total number of waiting vehicles on the lane. There are different definitions of a "waiting" state of a vehicle. In~\cite{wei2018intellilight}, a vehicle with a speed of less than 0.1 m/s is considered as waiting; in ~\cite{BSSN+05,KWBV08}, a vehicle without movement in terms of position is considered as waiting. 
    \item Waiting time. The waiting time of a vehicle is defined as the time period a vehicle has been in the ``waiting'' state (see queue length above for the definition of ``waiting''). 
    The definition on the beginning time of a waiting period may be different: in~\cite{wei2018intellilight,VaOl16}, they consider the waiting time starting from the last timestamp the vehicle moved, while~\cite{Wier00,BPT14,PBTB+13} consider the waiting time from the time the vehicle enters the road network. 
    \item Volume. Volume of a lane is defined as the number of vehicles on the lane, which equals to the sum of queuing vehicles and moving vehicles on the lane.
    \item Delay. Delay of a vehicle is defined as the time a vehicle has traveled within the environment minus the expected travel time (which equals the distance divided by the speed limit). 
    \item Speed. The speed of a vehicle is used to measure how fast the vehicle travels, which could be largely influenced by a pre-defined speed limit. Most literature uses a speed score, which is calculated by vehicle speed divided by the speed limit.
    \item Phase duration. Phase duration of the current phase is defined as how long the current phase has lasted.
    \item Congestion. Some studies take the congestion of the outgoing approach into account for effective learning for the cases of congestion and no congestion. The congestion of a lane can be defined either as an indicator (0 for no congestion and 1 for congestion) or the level of congestion, which equals to the number of vehicles divided by the maximum allowed vehicles on the lane.
    \item Positions of vehicles. The positions of vehicles are usually integrated as an image representation, which is defined as a matrix of vehicle positions, with '1' indicates the presence of a vehicle on a location, and '0' the absence of a vehicle on that location~\cite{VaOl16,mousavi2017traffic,wei2018intellilight}.
    \item Phase: The phase information is usually integrated into the state through an index of the current phase in the pre-defined signal phase groups\cite{EAA13,wei2018intellilight,VaOl16}.
\end{itemize}

There are also some variants of these elements in state representation. These elements can be defined on vehicle level as an image representation with the position of vehicles~\cite{VaOl16,wei2018intellilight}), on lane level by summing or averaging over all vehicles on corresponding lanes~\cite{wei2018intellilight}. 

Recently, there is a trend of using more complicated states in RL-based traffic signal control algorithms in the hope of gaining a more comprehensive description of the traffic situation. Specifically, recent studies propose to use images~\cite{VaOl16,mousavi2017traffic,wei2018intellilight} to represent the state, which results in a state representation with thousands or more dimensions. However, learning with such a high dimension for state often requires a huge number of training samples, meaning that it takes a long time to train the RL agent. And more importantly, longer learning schedule does not necessarily lead to significant performance gain, as the agent may have a more difficult time extracting useful information from the state representation.

\subsubsection{Reward Functions}

\begin{table}[htbp]
\caption{Factors in reward functions}
\label{tab:reward}
\begin{center}
\begin{tabular}{p{3cm} p{12cm}}
\toprule
Element  & References \\
\midrule
Queue length & \cite{abdoos2011traffic,AMB14,BGS10,SaCa10,CBYT11,MaDH16,SCGC08,wei2018intellilight,Wier00,PraB11,IKKK06,KWBV08,SSPN+05,WVVK04-2,WVVK04,VaOl16,KhGo12,KhGE12,aslani2017adaptive,aslani2018traffic,chu2019multi,lit2019,frep2019,colight2019, demolight,metaLight}\\ \midrule
Waiting time & \cite{MaDH16,wei2018intellilight,BWKG10,XXL13,BPT14,PBTB+13,PraB11,VaOl16,GCN18,chu2019multi}\\ \midrule
Change of delay & \cite{ElAb10,ElAb12,EAA13,ALUK10,mousavi2017traffic} \\ \midrule
Speed  &  \cite{wei2018intellilight,VaOl16,Casa17}\\ \midrule
Number of stops & \cite{VaOl16}\\ \midrule
Throughput & \cite{SCGC08,SaCa10,wei2018intellilight,XXL13,aslani2017adaptive,aslani2018traffic}\\  \midrule
Frequency of signal change & \cite{wei2018intellilight,VaOl16} \\ \midrule
Accident avoidance & \cite{VaOl16}\\ \midrule
Pressure & \cite{presslight2019, mpLight} \\
\bottomrule
\end{tabular}
\label{tab:reward-functions}
\end{center}
\end{table}

A reward defines the goal in a reinforcement learning problem. Equivalently, we can think of RL as an approach of optimization towards the objective, and this objective is specified as the reward function in the RL context. At each decision point, the agent takes action on the environment, and the environment sends a single numerical value called the reward to the agent. The agent's objective is to maximize the total reward it receives over the long run. 

In the traffic signal control problem, although the ultimate objective is to minimize the travel time of all vehicles, travel time is hard to serve as an effective reward in RL for several reasons. First, the travel time of a vehicle is influenced not only by the traffic signals, but also by other factors like the free-flow speed of a vehicle. Second, optimizing the travel time of all vehicles in the network becomes especially harder when the destination of a vehicle in unknown to the traffic signal controller in advance (which is often the case in the real world). Under such circumstances, the travel time of a vehicle can only be measured after it completes its trip when multiple actions have been taken by multiple intersections in the network. Therefore, the reward function is usually defined as a weighted sum of the factors in Table~\ref{tab:reward} that can be effectively measured after an action:

\begin{itemize}
    \item Queue length. The queue length is defined as the sum of queue length $L$ over all approaching lanes, where $L$ is calculated as the total number of waiting vehicles on the given lane. Similar to the definition of queue length in Section~\ref{sec:state}, there are different definitions on a "waiting" state of a vehicle. Minimizing the queue length is equivalent to minimizing total travel time.
    \item Waiting time. The waiting time of a vehicle is defined as the time a vehicle has been waiting. Similar to the definition of waiting time in Section~\ref{sec:state}, there are different definitions on how to calculate the waiting time of a vehicle. A typical reward function considers the negative value of the waiting time experienced by the vehicles.
    \item Change of delay. The change (saving) in the total cumulative delay is the difference between the total cumulative delays of two successive decision points. The total cumulative delay at time $t$ is the summation of the cumulative delay, up to time $t$, of all the vehicles that are currently in the system.  
    \item Speed. A typical reward takes the average speed of all vehicles in the road network. A higher average speed of vehicles in the road network indicates the vehicles travel to their destinations faster. 
    \item Number of stops. A reward can use the average number of stops of all vehicles in the network. Intuitively, the smaller the number of stops, the more smoothly the traffic moves.
    \item Throughput. The throughput is defined as the total number of vehicles that pass the intersection or leave the network during a certain time interval after the last action. Maximizing the throughput also helps to minimize the total travel time of all vehicles, especially when the road network is congested.
    \item Frequency of signal change. The frequency of signal change is defined as the number of times the signal changes during a certain time period. Intuitively, the learned policy should not lead to flickering, i.e., changing the traffic signal frequently, as the effective green time for vehicles to pass through the intersection might be reduced.
    \item Accident avoidance. There are also some studies that have special considerations for accident avoidance. For example, there should not be many emergency stops. Furthermore, jams or would-be collisions should be prevented.
    \item Pressure of the intersection. In~\cite{presslight2019}, the pressure of an intersection is defined as the sum of absolute pressure of every traffic movement. Intuitively, a higher pressure indicates a higher level of imbalances between the number of incoming lanes and outgoing lanes. 
\end{itemize}

There are also some variants of these factors to measure the immediate reward after an action. The reward could be defined as the values of the factors at certain decision points or defined as the difference between the corresponding total cumulative values over a certain period. Since most of these factors are a result of a sequence of actions and the effect of one action can hardly be reflected, whether to use these factors as the original value or as their difference still remains to be discussed.

Although defining reward as a weighted linear combination of several factors is a common practice in existing studies, there are two concerning issues with these ad-hoc designs in the traffic signal control context. First, there is no guarantee that maximizing the proposed reward is equivalent to optimizing travel time since they are not directly connected in transportation theory. Second, tuning the weights for each reward function component is rather tricky, and minor differences in the weight setting could lead to dramatically different results. Although the factors mentioned in Table~\ref{tab:reward} are all correlated with travel time, different weighted combinations of them yield very different results. Unfortunately, there is no principled way yet to select those weights.

\subsubsection{Action Definitions}
Now there are mainly four types of actions as shown in Table~\ref{tab:action-definitions}: 

\begin{itemize}
    \item Set current phase duration. Here, the agent learns to set the duration for the current phase by choosing from pre-defined candidate time periods. 
    \item Set cycle-based phase ratio. Here, the action is defined as the phase split ratio that the signal will set for the next cycle. Usually, the total cycle length is given, and the candidate phase ratio is pre-defined. 
    \item Keep or change the current phase in a cycle-based phase sequence. Here, an action is represented as a binary number, which indicates the agent decides to keep the current phase or change to the next phase.  
    \item Choose the next phase. Decide which phase to change to in a variable phase sequence, in which the phase sequence is not predetermined. Here, the action is the phase index that should be taken next. As a result, this kind of signal timing is more flexible, and the agent is learning to select a phase to change to without assumptions that the signal would change cyclically.
\end{itemize}

\begin{table}[htbp]
\caption{Action definitions}
\begin{center}
\begin{tabular}{p{3cm} p{12cm}}
\toprule
Action & References \\
\midrule
Set current phase duration & \cite{XXL13,aslani2017adaptive,aslani2018traffic} \\ \midrule
Set phase split & \cite{abdoos2011traffic,AMB14,BGS10,CBYT11,Casa17} \\ \midrule
Keep or change & \cite{MaDH16,wei2018intellilight,BPT14,PBTB+13,VaOl16} \\ \midrule
Choose next phase & \cite{SaCa10,ElAb10,ElAb12,EAA13,SCGC08,PraB11,Wier00,ALUK10,BWKG10,IKKK06,KWBV08,SSPN+05,WVVK04-2,WVVK04,KhGo12,KhGE12,mousavi2017traffic,GCN18,chu2019multi,lit2019,frep2019,presslight2019,colight2019, demolight,metaLight, mpLight} \\
\bottomrule
\end{tabular}
\label{tab:action-definitions}
\end{center}
\end{table}

\subsection{Learning Approaches}
There are varied algorithmic frameworks for RL methods from different perspectives. Depending on whether to learn the state-transition function or not, an RL method can be categorized as a model-based or model-free method, respectively. Depending on whether to learn the value function or to explicitly learn the policy parameter, an RL method can be categorized as a value-based or policy-gradient method, respectively (and the combination of these two is an actor-critic method). Depending on whether the functions, policies, and models are learned through tables with one entry for each state (or state-action pair) or through parameterized function representation, an RL method can be categorized as a tabular or approximation method, respectively.

\subsubsection{Model-based vs. model-free methods} 

\begin{table}[hbt]
\caption{Model-based and model-free methods in RL-based traffic signal control methods}
\label{tab:RL-model}
\begin{tabular}{p{3.7cm} p{7.3cm} p{4cm}}
\toprule
      & References  & Strengths\\ \midrule
Model-based methods     & \cite{SaCa10,Wier00,KWBV08,SSPN+05,WVVK04-2,WVVK04,KhGo12,KhGE12}  & Models the state transitions, explores the state space efficiently by planning and improves convergence speed.
\\  \midrule
Model-free methods & \cite{abdoos2011traffic,AMB14,BGS10,CBYT11,ElAb10,ElAb12,EAA13,MaDH16,SCGC08,wei2018intellilight,XXL13,BPT14,PBTB+13,PraB11,ALUK10,BWKG10,IKKK06,VaOl16,mousavi2017traffic,Casa17,aslani2017adaptive,aslani2018traffic,GCN18,chu2019multi,lit2019,frep2019,presslight2019,colight2019, demolight, metaLight, mpLight} &  No need to handcraft/pre-train transition models, learns the models directly with the policy.
\\ \bottomrule
\end{tabular}
\end{table}

Depending on the modeling philosophy of RL, literature~\cite{ADBB17, KLM96} divides current RL methods into two categories: model-based methods and model-free methods. For a problem with large state and action space, it is usually difficult to conduct millions of experiments to cover all the possible cases. Model-based methods try to model the transition probability among states explicitly (i.e., learning Equation~\eqref{eq:transition} or~\eqref{eq:transition-MARL}), which can be used to sample the environment more efficiently (agents can acquire data samples according to this transition probability). Concretely, given this model, we will know which state each action will take the agent to. In contrast, model-free methods directly estimate the reward for state-action pairs and choose the action based on this. Hence, model-free methods can be used even the transition probability is hard to model.

In the context of traffic signal control, to develop a model-based model, it requires the transition probability of the environment to be known or modeled, like the position, speed, and acceleration of all vehicles, and the operations of all the traffic signals. However, since people's driving behavior is different and hard to predict. Currently, most RL-based methods for traffic signal control are model-free methods, as is shown in Table~\ref{tab:RL-model}.

\subsubsection{Value-based, policy-based vs. actor-critic methods.}

Depending on the different ways of estimating the potential reward and select action, reinforcement learning methods can be categorized into the following three categories, as is shown in summarized in Table~\ref{tab:RL-value-policy}: 

\begin{table}[hbt]
\caption{Value- and policy-based methods}
\label{tab:RL-value-policy}
\begin{tabular}{p{2.5cm} p{8.5cm} p{4.0cm}}
\toprule
                & References  &  Strengths\\ \midrule
Value-based     & \cite{abdoos2011traffic,AMB14,BGS10,SaCa10,CBYT11,ElAb10,ElAb12,EAA13,MaDH16,SCGC08,wei2018intellilight,XXL13,BPT14,PBTB+13,Wier00,ALUK10,BWKG10,IKKK06,KWBV08,SSPN+05,WVVK04-2,WVVK04,VaOl16,KhGo12,KhGE12,GCN18,lit2019,frep2019,presslight2019,colight2019,metaLight,mpLight}  &  Combines policy evaluation (predicting the value of a policy) and control (finding the best policy), easy to be understood through interpreting the predicted values  \\ \midrule
Policy-based &  \cite{PraB11,genders2016using,mousavi2017traffic,Casa17,aslani2017adaptive,aslani2018traffic,chu2019multi,demolight,rizzo2019time}  &   Directly learns the policy and optimizes with faster convergence.  \\ 
\bottomrule
\end{tabular}
\end{table}

\begin{itemize}
\item Value-based methods approximate the state-value function or state-action value function (i.e., how rewarding each state is or state-action pair is), and the policy is implicitly obtained from the learned value function. Value-based methods, including Q-learning~\cite{AMB11, CBYT11}, and DQN~\cite{VaOl16, wei2018intellilight}, directly model the state-values or state action values (e.g., under the current traffic situation, how much average speed increase/decrease will take into effect if one action is conducted). In this way, the state and reward can be directly fed into the model without extra processing. However, these methods are usually combined with an $\epsilon$-greedy action selection methods and hence will result in a nearly deterministic policy when $\epsilon$ finally decays to a small number (i.e., it is deterministic which action will be conducted under certain states). This may cause the agent to stuck in some unseen or ill-represented cases without improving. In addition, these methods can only deal with discrete actions because it requires a separate modeling process for each action.
\item Policy-based methods directly update the policy (e.g., a vector of probabilities to conduct actions under specific state) parameters along the direction to maximize a pre-defined objective (e.g., average expected reward). Policy-based methods, try to learn a probability distribution of different actions under a certain state. The advantage of policy-based methods is that it does not require the action to be discrete. Besides, it can learn a stochastic policy and keep exploring potentially more rewarding actions.
Actor-Critic is one of the widely used methods in policy-based methods. It includes the value-based idea in learning the policy for the action probability distribution, with an actor controls how our agent behaves (policy-based), and the critic measures how good the conducted action is (value-based). Actor-Critic methods in traffic signal control~\cite{mousavi2017traffic, PraB11, aslani2017adaptive, aslani2018developing} utilize the strengths of both value function approximation and policy optimization, have shown excellent performance in traffic signal control problems. 
\end{itemize}

\subsubsection{Tabular methods and approximation methods} 

For small-scale discrete reinforcement learning problem in which the state-action pairs can be enumerated, it is a common practice to use a table to record the value functions, policies, or models. However, for a large-scale or continuous problem (state or action is continuous), it is not realistic to enumerate all the state-action pairs. In this case, we need to estimate the value function, policy, or model using an approximation function of the state and action.

\begin{table}[hbt]
\caption{Tabular methods and approximation methods}
\label{tab:RL-approx}
\begin{tabular}{p{2.8cm} p{8.2cm} p{4cm}}
\toprule
         & References  & Strengths \\ \midrule
Tabular methods                                                 & \cite{abdoos2011traffic,BGS10,SaCa10,CBYT11,ElAb10,ElAb12,EAA13,SCGC08,XXL13,BPT14,PBTB+13,Wier00,PraB11,IKKK06,KWBV08,SSPN+05,WVVK04-2,WVVK04,KhGo12,KhGE12} &  Efficient to search the optimal policy \\  \midrule
Approximation methods & \cite{AMB14,MaDH16,wei2018intellilight,ALUK10,BWKG10,VaOl16,mousavi2017traffic,Casa17,aslani2017adaptive,aslani2018traffic,GCN18,chu2019multi,lit2019,frep2019,presslight2019,colight2019, demolight, metaLight, mpLight,genders2016using} & Handles the high dimension of state/action space and generalizes well to unexplored situations.
\\ \bottomrule
\end{tabular}
\end{table}

In the problem scenario of signal control, earlier methods use simple state features like levels of congestion (which can be converted to discrete state values) and employ tabular Q-learning \cite{APG03,ElAb10} to learn the value function. These methods thus, cannot scale up to a large state space for complicated scenarios in traffic signal control. Also, tabular methods will treat samples with similar features as two completely different states, which will decrease the efficiency of utilizing samples in training. To solve these problems, recent studies propose to apply function approximation using continuous state representation\cite{BPT14,ADBB17,wei2018intellilight,MaDH16,VaOl16,AMB14,BaKl14,EAA13,ElAb12,KhGo12,ALUK10}. The function approximation could be achieved by multiple machine learning techniques, like tile coding~\cite{albus1971theory} or deep neural networks. These methods can utilize the samples with similar state values more efficiently and deal with more informative features with continuous ranges, e.g., the position of vehicles. The categorization of RL-based traffic signal control methods in terms of tabular or approximation is summarized in Table~\ref{tab:RL-approx}.

\subsection{Coordination Strategies}
Coordination could benefit signal control for multi-intersection scenarios. Since recent advances in RL improve the performance on isolated traffic signal control~\cite{VaOl16,wei2018intellilight}, efforts have been performed to design strategies that cooperate with MARL agents. Liturature~\cite{ClBo98} categorizes MARL into two classes: Joint Action Learners and Independent Learners. Here we extend this categorization for the traffic signal control problem, as is shown in Table~\ref{tab:coordination}.  

\begin{itemize} 

\item Global single agent. A straightforward solution is to use one central agent to control all the intersections~\cite{PraB11}. It directly takes the state as input and learns to set the joint actions of all intersection at the same time. However, this method can result in the curse of dimensionality, which encompasses the exponential growth of the state-action space in the number of state and action dimensions. 

\item Joint action modeling. \cite{KWBV08} and \cite{VaOl16} consider explicit coordination mechanisms between learning agents using coordination graphs, extending \cite{Wier00} using max-plus algorithm. They factorize the global Q-function as a linear combination of local subproblems: $\hat{Q}(o_1,\dots,o_N,\pmb{a})=\Sigma_{i,j} Q_{i,j} (o_i,o_j, \action_i, \action_j)$, where $i$ and $j$ corresponds to the index of neighboring agents.

\item Independent RL. There are also a line of studies that use individual RL agents to control the traffic signals in the multi-intersection network~\cite{abdoos2011traffic,BGS10,SaCa10,ElAb10,MaDH16,SCGC08,BPT14,PBTB+13}. In these methods, each intersection is controlled by an RL agent which senses part of the environment and adapt (or react) to it accordingly and eventually form several subgroups of synchronization. 

\begin{table}[t!]
\caption{Different coordination methods for traffic signal control}
\label{tab:coordination}
\begin{tabular}{p{3.8cm} p{4.0cm} p{6.2cm}}
\toprule
Coordination Strategies &  Objective \& Explanation  & References \\ \midrule
Global single agent & $max_{\pmb{a}}Q(s,\pmb{a})$, where $s$ is the global environment state, $\pmb{a}$ is the joint action of all intersections  & \cite{PraB11,Casa17}  \\ \midrule
Joint action modeling & $max_{a_i,a_j} \Sigma_{i,j} Q_{i,j} (o_i,o_j, \action_i, \action_j)$, where $o_i$ and $o_j$ are the observation of two neighboring agents $i$ and $j$ & \cite{ElAb12,EAA13,XXL13,KWBV08,VaOl16}  \\ \midrule
Independent RL without communication & $ max_{\action_i}\Sigma_i Q_i(o_i, \action_i)$, where $o_i$ is the local observation of intersection $i$, $\action_i$ is the action of intersection $i$&  \cite{abdoos2011traffic,BGS10,SaCa10,MaDH16,SCGC08,BPT14,PBTB+13,Wier00,IKKK06,SSPN+05,WVVK04-2,WVVK04,KhGo12,KhGE12,aslani2017adaptive,aslani2018traffic,chu2019multi,lit2019,frep2019,presslight2019, demolight, metaLight, mpLight}    \\ \midrule
Independent RL with communication & $max_{\action_i}\Sigma_i Q_i(\Omega(o_i,\mathcal{N}_i), \action_i)$, where $\mathcal{N}_i$ is the neighborhood representation of intersection $i$, $\Omega(o_i,\mathcal{N}_i)$ is the function that models local observations and the observations of neighborhoods. & \cite{ALUK10,ElAb10,GCN18,colight2019,zhang2019integrating}  \\ 
\bottomrule
\end{tabular}
\end{table}

\begin{itemize}
    \item Without communication. These approaches do not use explicit communication to resolve conflicts. Instead, the observation of agent $i$ is defined on the local traffic condition of intersection $i$. In some simple scenarios like arterial networks, this approach has performed well with the formation of several mini green waves.  However, when the environment becomes complicated, the non-stationary impacts from neighboring agents will be brought into the environment, and the learning process usually cannot converge to stationary policies if there are no communication or coordination mechanisms among agents~\cite{NoVr12}.  In highly dynamic environments, an agent might not have the time to perceive a change and adapt to it before the environment has already changed again.
    \item With communication. While using individual RL to control each agent in a multi-agent system can form coordination under simple environments, communication between agents can enable agents to behave as a group, rather than a collection of individuals in complex tasks where the environment is dynamic and each agent has limited capabilities and visibility of the world~\cite{SF16}. 
    It is especially important for traffic signal control in a real-world scenario where the intersections are in close proximity, and the traffic is highly dynamic. While some studies also add neighbor's traffic condition directly into $o_i$, \cite{GCN18,colight2019} proposes to use Graph Convolutional Network~\cite{SKBV+18} where cooperating agents learn to communicate amongst themselves. This method not only learns the interactions between the hidden state of neighboring agents and the target agent but also learn multi-hop influences between intersections.
\end{itemize}

\end{itemize}

\subsection{Experimental Settings}
In this section, we will introduce some experimental settings that will influence the performance of traffic signal control strategies: simulation environment, road network setting, and traffic flow setting.
\begin{itemize}
    \item Simulation environment. Deploying and testing traffic signal control strategies involves high cost and intensive labor. Hence, simulation is a useful alternative before actual implementation. Simulations of traffic signal control often involve large and heterogeneous scenarios, which should account for some specific mobility models in a vehicular environment, including car following, lane changing, and routing. Since mobility models can significantly affect simulation results, the simulated model must be as close to reality as possible.
    \item Road network. Different kinds of the road network are explored in the current literature, including synthetic and real-world road network. While most studies conduct experiments on the synthetic grid network, the scale of the network is still relatively small compared to the scale of a city.
    \item Traffic flow. Traffic flow in the simulation can also influence the performance of control strategies. Usually, the more dynamic and heavier the traffic flow is, the harder for an RL method to learn an optimal policy.
\end{itemize}

\subsubsection{Simulation environment}
Various publicly available traffic simulators are currently in use by the research community. In this section, we briefly introduce some open-source tools used in the current traffic signal control literature. Specifically, since RL-based methods require detailed state representation like vehicle-level information, most literature relies on microscopic simulation, in which movements of individual vehicles are represented through microscopic properties such as the position and velocity of each vehicle. Other proprietary simulators like Paramics~\footnote{https://www.paramics-online.com/} or Aimsun~\footnote{https://www.aimsun.com} will not be introduced here. For a detailed comparison of the open-source simulators, please refer to~\cite{simulator11}.

\begin{table}[hbt]
\caption{Different simulators used in literature for RL-based traffic signal control }
\label{tab:simulators}
\begin{tabular}{p{1.3 cm} p{2.5cm} p{6.0cm} p{5.2cm}}
\toprule
Name & Latest update & References  & Strengths\\ \midrule
GLD &   1.0 (Jan. 2005)  &  Strengths
\cite{Wier00,PraB11,BWKG10,IKKK06,KWBV08,SSPN+05,WVVK04-2,WVVK04,KhGo12,KhGE12}
 &  Co-learning of vehicle navigation and traffic signal control.
 \\ \midrule
SUMO &  1.1.0 (Dec. 2018) & 
\cite{MaDH16, wei2018intellilight,VaOl16,mousavi2017traffic,GCN18,chu2019multi} & Real-time visualization, various vehicle behaviour models, supportive built-on computational frameworks for RL.
\\ \midrule
AIM  &  1.0.4 (Mar. 2017)  & 
\cite{PBTB+13, BPT14}  & Mixed (automated and human-like) vehicle simulation.\\ \midrule
CityFlow &  0.1 (Mar. 2017)   &  \cite{presslight2019, colight2019, demolight, metaLight, mpLight,lit2019,frep2019}  & Multi-thread simulation for large scale traffic settings and multi-agent reinforcement learning, driving behavior modelling.
\\ \midrule
Others & \multicolumn{3}{l}{ \begin{tabular}[c]{@{}l@{}}  Paramics~\cite{BGS10,ElAb10,ElAb12,EAA13,XXL13}, \\
Aimsun~\cite{abdoos2011traffic,AMB14,Casa17,aslani2017adaptive,aslani2018traffic}, \\
Matlab~\cite{ALUK10}, QLTST~\cite{CBYT11}, UTC~\cite{SCGC08}
\end{tabular} } \\
\bottomrule
\end{tabular}
\end{table}

\begin{itemize}
    \item The Green Light District (GLD)~\footnote{https://sourceforge.net/projects/stoplicht/}. GLD is an open-source traffic flow simulator that can be used to evaluate the performance of AI traffic light controller algorithms and co-learning navigation algorithms for cars. Vehicles enter the network at edge nodes (i.e., one end of roads which is not connected with an intersection), and each edge node has a certain probability of generating a vehicle at each timestep (spawn rate). Each generated vehicle is assigned to one of the other edge nodes as a destination. The distribution of destinations for each edge node can be adjusted. There can be several types of vehicles, defined by their speed, length, and number of passengers.
    \item The Autonomous Intersection Management (AIM)~\footnote{http://www.cs.utexas.edu/\~aim/}. AIM is a microscopic traffic simulator which mainly supports a Manhattan topology of North-South and East-West multi-lane roads joined by many intersections. Developed by the Learning Agents Research Group at the University of Texas at Austin, AIM supports vehicles to navigate, accelerate, and decelerate, as well as subtle details, including variable vehicle sizes.
    \item Simulation of Urban MObility (SUMO)~\footnote{http://sumo.sourceforge.net}. SUMO is an open-source program for traffic simulation, mainly developed by the Institute of Transportation Systems, located at German Aerospace Center. Among other features, it allows the existence of different types of vehicles, roads with several lanes, traffic lights, graphical interface to view the network and simulated entities, and interoperability with other applications at run-time through an API called TraCI. Moreover, the tool can be accelerated by allowing a version without a graphical interface. Also, it will enable importing real-world road networks from OpenStreetMap, and the vehicles can enter the system from any position in the network. SUMO also supports other upper-level computational frameworks for deep RL and control experiments for traffic microsimulation.
    \item CityFlow.  CityFlow~\cite{zhang2019cityflow} is a multi-agent reinforcement learning environment for large scale city traffic scenarios. It supports real-world definitions for the road network and traffic flow and is capable of multi-thread simulation for city-wide traffic settings. It also provides APIs for reinforcement learning, which is suitable for tasks like traffic signal control and driving behavior modeling.

\end{itemize}

\subsubsection{Road network}

\begin{table}[htb]
\caption{Different road network in literature for RL-based traffic signal control}
\label{tab:exp-network}
\begin{tabular}{p{1.5cm} p{1.5cm} p{8.5cm}}
\toprule
                   & Intersections   & References \\ \midrule
Synthetic network & $<10$     & \begin{tabular}[c]{@{}l@{}}                        \cite{AMB14,MaDH16,XXL13} \\
                \cite{BPT14,PBTB+13,Wier00,GCN18} \\
                \cite{PraB11,ALUK10,VaOl16}\\
                \end{tabular}        \\ \cmidrule{2-3}
                   & $10-20$   & \begin{tabular}[c]{@{}l@{}} \cite{SaCa10,BWKG10,IKKK06,KWBV08} \\
                   \cite{SSPN+05,WVVK04-2,WVVK04,KhGo12}\\
                   \cite{KhGE12}         \end{tabular}   \\ \cmidrule{2-3}
                   & $>=20$    &  
                   \begin{tabular}[c]{@{}l@{}} \cite{chu2019multi}(25 intersections),\cite{abdoos2011traffic} (50 intersections) \\
                   ~\cite{colight2019}(100)
                \end{tabular}
                \\ \midrule
Real road network  & \multicolumn{2}{l}{
                            \begin{tabular}[c]{@{}l@{}}
                        \cite{BGS10}(29 intersections),~\cite{Casa17}(43 intersections),\\
                        \cite{aslani2017adaptive,aslani2018traffic}(50 intersections),~\cite{SCGC08}(64 intersections), \\
                        \cite{ElAb12,EAA13}(59 intersections),\\
                        ~\cite{lit2019}(16 intersections),~\cite{frep2019}(5 intersections) \\ ~\cite{presslight2019}(16 intersections), ~\cite{colight2019}(196 intersections), \\
                        ~\cite{mpLight} (2150 intersections)
                        \end{tabular} 
                        }
                        \\ 
                   \bottomrule        
\end{tabular}
\end{table}

At a coarse scale, a road network is a directed graph with nodes and edges representing intersections and roads, respectively. Specifically, a real-world road network can be more complicated than the synthetic network in the road properties (like the position, shape, and speed limit of every lane), intersections, traffic movements, traffic signal logic (movement signal phases). Table~\ref{tab:exp-network} summarizes the road network in literature. While existing studies conduct experiments on different kinds of road networks, the number of intersections in the tested network is mostly smaller than twenty. 

\subsubsection{Traffic flow}
In the traffic flow dataset, each vehicle is described as $(o,t,d)$, where $o$ is the origin location, $t$ is time, and $d$ is the destination location. Locations $o$ and $d$ are both locations on the road network. Table~\ref{tab:exp-traffic} summarizes the traffic flow utilized in the current studies. In the synthetic traffic flow data, either $o$, $t$ or $d$ could be synthesized to generate uniform or dynamic changing flow in different levels of traffic.

\begin{table}[htb]
\caption{Traffic flow used in literature for RL-based traffic signal control. Traffic with arrival rate less than 500 vehicles/hour/lane is considered as light traffic in this survey, otherwise considered as heavy.}
\label{tab:exp-traffic}
\begin{tabular}{p{1.5cm} p{1.0cm}p{1.5cm} p{11 cm}}
\toprule
                &                &          & References \\ \midrule
Synthetic data  & light          & uniform         &  \begin{tabular}[c]{@{}l@{}}
                                                        \cite{wei2018intellilight,XXL13,Wier00,mousavi2017traffic}  
                                                    \end{tabular} \\ \cmidrule{3-4}
                &                & dynamic       & \begin{tabular}[c]{@{}l@{}} \cite{abdoos2011traffic,AMB14,CBYT11,MaDH16} \\
                \cite{SCGC08,ALUK10,BWKG10,KWBV08}\\
                \cite{VaOl16,GCN18}  
                \end{tabular}
                \\ \cmidrule{2-4}
                & heavy          & uniform         &  \begin{tabular}[c]{@{}l@{}}
                                                        \cite{BPT14,PBTB+13,PraB11} \\
                                    \cite{SSPN+05,WVVK04-2,WVVK04,wei2018intellilight}  
                                                    \end{tabular} \\ \cmidrule{3-4}
                &                & dynamic       &   \begin{tabular}[c]{@{}l@{}} \cite{abdoos2011traffic,BGS10,wei2018intellilight,ALUK10} \\
                \cite{,BWKG10,KhGo12,KhGE12}  \\
                \cite{SSPN+05,chu2019multi,lit2019,frep2019}
\end{tabular}      \\ \midrule
Real data & \multicolumn{3}{c}{
\begin{tabular}[c]{@{}l@{}}
\cite{ElAb10,ElAb12,EAA13,wei2018intellilight,Casa17}\\
\cite{aslani2017adaptive,aslani2018traffic, lit2019, frep2019}\\
\cite{presslight2019,colight2019,  demolight, metaLight, mpLight}
\end{tabular}} \\
\bottomrule
\end{tabular}
\end{table}

\subsection{Challenges in RL for Traffic Signal Control}

Current RL-based methods have the following challenges: 

\subsubsection{Design of RL formulation}
A key question for RL is how to formulate the RL setting, i.e., the reward and state definition. In existing studies~\cite{Wier00,VaOl16,wei2018intellilight}, a typical reward definition for traffic signal control is a weighted linear combination of several components such as queue length, waiting time and delay. The state features include components such as queue length, number of cars, waiting time, and current traffic signal phase. In recent work~\cite{VaOl16,wei2018intellilight}, images of vehicles' positions on the roads are also considered in the state.  

However, all of the existing work take an ad-hoc approach to define reward and state. Such an ad-hoc approach will cause several problems that hinder the application of RL in the real world. First, the engineering details in formulating the reward function and state feature could significantly affect the results. For example, if the reward is defined as a weighted linear combination of several terms, the weight on each term is tricky to set, and a minor difference in weight setting could lead to dramatically different results. Second, the state representation could be in a high-dimensional space, especially when using traffic images as part of the state representation~\cite{VaOl16,wei2018intellilight}. With such a high-dimensional state representation, the neural network will need much more training data samples to learn and may not even converge. Third, there is no connection between existing RL approaches and transportation methods. Without the support of transportation theory, it is highly risky to apply these purely data-driven RL-based approaches in the real physical world. 

\paragraph{Learning efficiency}
While learning from trial-and-error is the key idea in RL, the learning cost of RL could be unacceptable for complicated problems. 
\nop{Although RL algorithms are useful in learning good solutions when the model of the environment is unknown in advance~\cite{SuBa98}; the solutions may only be achieved after an extensive number of trial-and-error, which is usually time-consuming.
}
Existing RL methods for games (e.g., Go or Atari games) usually require a massive number of update iterations of RL models to yield impressive results in simulated environments. These trial-and-error attempts will lead to real traffic jams in the traffic signal control problem. Therefore, how to learn efficiently (e.g., learning from limited data samples, efficient exploration, transferring learned knowledge) is a critical question for the application of RL in traffic signal control.

\paragraph{Credit assignment}
The credit assignment problem is one of the heatedly investigated problems in RL, which considers the distribution of credits for success (or blame for failure) of an action~\cite{Sutt84}. In the traffic signal control problem, the traffic condition is a consequence of several actions that traffic signal controllers have been taken, which brings two problems: (1) one action may still have effect after several steps of actions; (2) the reward at each timestamp is a result of the combination of consequent actions from several agents. Unlike in Atari or Go games, where people sometimes assign the final score of an episode to all the actions associated with this episode, the action in the traffic signal control problem may have an affecting time interval that may be dynamically changing and needs to be further investigated.

\paragraph{Safety issue}
Making reinforcement learning agents acceptably safe in physical environments is another important area for future research. While RL methods learn from trial-and-error, the learning cost of RL could be critical or even fatal in the real world as the malfunction of traffic signals might lead to accidents. Therefore, how to adopt risk management into RL helps prevent unwanted behavior during and after the learning process of RL.

%% file: conclusion.tex

\section{Conclusion}
\label{sec:conclusion}

In this article, we present an overview of traffic signal control. We first introduce some terms and objectives in traffic signal control problem. Then we introduce some of the classical transportation approaches. Next we review state-of-the-art RL-based traffic signal control methods from the perspective of the formulation of RL agent (state, reward, action and ways of coordination) and their experiment settings. After that, we briefly discuss some challenges for future research directions on RL-based traffic signal control methods. Hopefully, this survey can provide a comprehensive view from both traditional transportation methods and reinforcement learning methods and can bring the traffic signal control research into a new frontier.

%% file: appendix.tex

\appendix
\section{Appendix A}

\subsection{Summary of RL methods}
RL-based traffic signal control methods are summarised in the Table~\ref{tab:overall-RL}.

\newpage
\begin{landscape}

\begin{table}[h]
\caption{Overall comparison of RL-based traffic signal control methods investigated in this survey}
\label{tab:overall-RL}
\begin{tabular}{lllllll p{1cm} p{1cm} p{1.2cm} p{1cm}}
\toprule
Paper name           & State     & Reward    & Action & Model & Method & Approx & Coope- ration & Simula- tor & Road net & Traffic flow \\ \midrule
\cite{abdoos2011traffic}   & 1         & 1         & 1      & 2     & 1      & 1      & 2           & 6          & 3(50)    & 2,4          \\
\cite{AMB14}               & 1         & 1         & 1      & 2     & 1      & 2      & 1           & 6          & 1(9)     & 2            \\
\cite{BGS10}               & 1,3       & 1         & 1      & 2     & 1      & 1      & 2           & 5          & 4(29)    & 4            \\
\cite{SaCa10}              & 3         & 1,7       & 3      & 1     & 1      & 1      & 2           & 1          & 2(15)    & uk           \\
\cite{CBYT11}              & 1         & 1         & 1      & 2     & 1      & 1      & -           & 7          & 5        & 2            \\
\cite{ElAb10}              & 1,3,4     & 3         & 3      & 2     & 1      & 1      & 3           & 5          & 5        & 5            \\
\cite{ElAb12}              & 1         & 3         & 3      & 2     & 1      & 1      & 4           & 5          & 4(59)    & 5            \\
\cite{EAA13}               & 1,6,9     & 3         & 3      & 2     & 1      & 1      & 4           & 5          & 4(59)    & 5            \\
\cite{MaDH16}              & 1,6,9     & 1,2       & 2      & 2     & 1      & 2      & 2           & 2          & 1(9)     & 2            \\
\cite{SCGC08}              & 1,9       & 1,7       & 3      & 2     & 1      & 1      & 2           & 8          & 4(64)    & 2            \\
\cite{wei2018intellilight} & 1,2,3,8,9 & 1,2,5,7,8 & 2      & 2     & 1      & 2      & -           & 2          & 5        & 1,3,4,5      \\
\cite{XXL13}               & 1         & 2,7       & 4      & 2     & 1      & 1      & 4           & 4          & 1(5)     & 1            \\
\cite{BPT14}               & 1,6       & 2         & 2      & 2     & 1      & 1      & 2           & 3          & 1(4)     & 3            \\
\cite{PBTB+13}             & 1,6       & 2         & 2      & 2     & 1      & 1      & 2           & 3          & 1(4)     & 3            \\
\cite{Wier00}              & 8         & 2         & 3      & 1     & 1      & 1      & 2           & 1          & 1(6)     & 1            \\
\cite{PraB11}              & 1,6       & 1,2       & 3      & 2     & 2    & 1      & 1           & 1          & 1(5)     & 3            \\
\cite{ALUK10}              & 5         & 3         & 3      & 2     & 1    & 2      & 3           & 4          & 1(5)     & 2,4          \\
\cite{BWKG10}              & 7,8       & 2         & 3      & 2     & 1      & 2      & 2,4           & 1          & 2(15)    & 2,4          \\
\cite{IKKK06}              & 7,8,10    & 2         & 3      & 2     & 1      & 1      & 2           & 1          & 2(15)    & uk           \\
\cite{KWBV08}              & 8         & 2         & 3      & 1     & 1      & 1      & 4           & 1          & 2(15)    & 2            \\
\cite{WVVK04-2}            & 8         & 2         & 3      & 1     & 1      & 1      & 2           & 1          & 2(15)    & 3            \\
\cite{WVVK04}              & 8         & 2         & 3      & 1     & 1      & 1      & 2           & 1          & 2(15)    & 3            \\
\cite{VaOl16}              & 8         & 1,2,5,6,8 & 2      & 2     & 1      & 2      & 4           & 2          & 1(4)     & 2            \\
\cite{KhGo12}              & 8         & 1         & 3      & 1     & 1      & 1      & 2           & 1          & 2(12)    & 4            \\
\cite{KhGE12}              & 8         & 1         & 3      & 1     & 1      & 1      & 2           & 1          & 2(12)    & 4            \\
\cite{mousavi2017traffic}  & 8         & 3         & 3      & 2     & 2    & 2      & -           & 2          & 5        & 1            \\
\cite{Casa17}              & 3,4       & 5         & 1      & 2     & 2      & 2      & 1           & 6          & 4(43)    & 5            \\
\cite{aslani2017adaptive}  & 1,3,9     & 1,7       & 4      & 2     & 2      & 2      & 2           & 6          & 4(50)    & 5            \\
\cite{aslani2018traffic}   & 1,3,9     & 1,7       & 4      & 2     & 2      & 2      & 2           & 6          & 4(50)    & 5            \\
\cite{GCN18}               & 1,4       & 2         & 3      & 2     & 1      & 2      & 3           & 2          & 1(6)     & 2            \\
\cite{chu2019multi}        & 1,2       & 1,2       & 3      & 2     & 2      & 2      & 2           & 2          & 3(25)    & 4            \\ 
\cite{lit2019}             & 1,9       & 1         & 3      & 2     & 1      & 2      & 2           & 2          & 4(16)    & 4            \\ 
\cite{frep2019}             & 1,9      & 1         & 3      & 2     & 1      & 2      & 2           & 2          & 4(5)    & 4            \\ 
\cite{presslight2019}       & 3,9     & 9       & 3      & 2     & 1      & 2      & 2           & 9          & 4(16)    & 5            \\ 
\cite{colight2019}       & 1,9     & 1       & 3      & 2     & 1      & 2      & 3           & 2          & 4(48)    & 5            \\ 
\bottomrule
\end{tabular}

\end{table}

\end{landscape}

\newpage

\begin{landscape}

\begin{table}[h]
\caption{Overall comparison of RL-based traffic signal control methods investigated in this survey}
\label{tab:overall-RL-1}
\begin{tabular}{lllllll p{1cm} p{1cm} p{1.2cm} p{1cm}}
\toprule
Paper name           & State     & Reward    & Action & Model & Method & Approx & Coope- ration & Simula- tor & Road net & Traffic flow \\ \midrule
\cite{metaLight}             & 1,9       & 1         & 3      & 2     & 1      & 2      & -           & 9          & 4(16)    & 4            \\ 
\cite{demolight}             & 1,9      & 1         & 3      & 2     & 1      & 2      & -           & 9          & 4(5)    & 4            \\ 
\cite{mpLight}       & 3,9     & 9       & 3      & 2     & 1      & 2      & 2           & 9          & 4(16)    & 5            \\ 
\cite{rizzo2019time}       & 2     & 7       & 3      & 2     & 1      & 2      & 3           & 2          & 4(1)    & 5            \\ 
\cite{wang2019stmarl}       & 1,2     & 1       & 2      & 2     & 1      & 2      & 3           & 4          & 4(4)    & 5            \\ 
\cite{zhang2019integrating}       & 1,2,3     & 1,2,5,7,8       & 2      & 2     & 1      & 2      & 3           & 2          & 3(36)    & 4            \\  
\bottomrule
\end{tabular}

\end{table}

\textbf{State}:  1. Queue; 2. Waiting time; 3. Volume; 4. Speed; 5. Delay; 6. Elapsed time; 7. Congestion; 8. Position of vehicles; 9. Phase; 10. Accident; 11. Pressure
\\
\textbf{Reward}: 1.Queue; 2. Waiting time; 3.Delay; 4. Accident; 5. Speed; 6. Number of stops; 7. Throughput 8. Frequency of signal change; 9. Pressure
\\ 
\textbf{Action}: 1. Phase split; 2. phase switch; 3. phase itself; 4 phase duration.  \\
\textbf{Model}:  1. Model-based; 2.Model-free. \\
\textbf{Method}: 1. Value-based; 2. Policy-based \\
\textbf{Approx}: 1. Tab; 2. Approx. \\
\textbf{Cooperation}: 1. Single; 2. IRL wo/ communication; 3. IRL w/ communication;  4. Joint action. \\
\textbf{Simulation}: 1. GLD; 2. SUMO; 3. AIM; 4. Matlab; 5. Paramics; 6. Aimsun; 7. QLTST; 8. UTC  9. CityFlow  \\
\textbf{Road net}: 1. Synthetic <10; 2. Synthetic 11-20; 3. Synthetic \textgreater{}=21; 4. Real 5. Single intersection                                       \\ 
\textbf{Traffic flow}:  1. Synthetic light uniform; 2. Synthetic light dynamic; 3. Synthetic heavy uniform; 4. Synthetic heavy dynamic; 5. Real-world data  

\end{landscape}
\newpage

%% file: ack.tex

\begin{acks}
This is acknowledgement.
\end{acks}